# A Tumor-Aware DenseNet–Swin Hybrid Learning with Boosted and Hierarchical Feature Spaces for Large-Scale Brain MRI Classification

Muhammad Ali Shah[1], Muhammad Mansoor Alam[1,2], Saddam Hussain Khan[3*]

[1]Riphah International University, Islamabad, Pakistan

[2*]Multimedia University, Malaysia

[3*] University of Engineering and Applied Sciences, Swat, Kanju Township, Pakistan

Correspondence Email: hengrshkhan822@gmail.com

## Abstract

According to the American Society of Clinical Oncology, the tenth leading cause of mortality is brain tumors, with the range of 40,000 to 80,000 deaths per year estimated to be caused by diagnostic errors. Radiologists tend to suffer fatigue, cognitive bias, and overwork, which may lead to misdiagnosis. Thus, computerized systems of diagnosis are necessary to assist radiologists and enhance the accuracy of diagnosis, but face challenges of tumor texture, contrast, and morphology variation. This study proposes an efficient Densely–Swin Hybrid (EDSH) framework for brain tumor MRI analysis, designed to jointly capture fine-grained texture patterns and long-range contextual dependencies. Two tumor-aware experimental setups are introduced to address class-specific diagnostic challenges. The first setup employs a Boosted Feature Space (BFS), where independently customized DenseNet and Swin_t branches learn complementary local and global representations that are dimension-aligned, fused, and boosted, enabling highly sensitive detection of diffuse glioma patterns by successfully learning the features of irregular shape, poorly defined mass, and heterogeneous texture. The second setup adopts a hierarchical DenseNet–Swin_t architecture with Deep Feature Extraction have Dual Residual connections (DFE+DR), in which DenseNet serves as a stem CNN for structured local feature learning, while Swin_t models global tumor morphology, effectively suppressing false negatives in meningioma and pituitary tumor classification by learning the features of well defined mass, location (outside brain) and enlargments in tumors (dural tail or upward extension). DenseNet is customized at the input level to match MRI spatial characteristics, leveraging dense residual connectivity to preserve texture information and mitigate vanishing-gradient effects. In parallel, Swin_t is tailored through task-aligned patch embedding and shifted-window self-attention to efficiently capture hierarchical global dependencies. At the target level, task-specific outputs from BFS and DFE+DR are systematically integrated in the SI block to enhance class separability and clinical reliability across heterogeneous tumor morphologies. Extensive evaluation on a large-scale MRI dataset (stringent 40,260 images across four tumor classes) demonstrates consistent superiority over standalone CNNs, Vision Transformers, and hybrids, achieving 98.50% accuracy and recall on the test unseen dataset. The further generalization of the proposed EDSH framework has been tested on an anonymous dataset, achieving 98.80% accuracy and 98.78% recall, highlighting strong potential for reliable clinical decision support.

**Keywords:** Deep Learning, Transformer, Swin, Boosting, ViT, CNN, Brain Tumor, Detection, MRI

1. **Introduction**

A tumor is associated with the uncontrolled and abnormal growth of cells in the human body, but not every tumor is cancer [1]. The brain, being the main organ of the human body, controls many important functions of the day. With the growth of a tumor in the brain, the growth, which may be irregular and invasive, may greatly hamper both physical and cognitive functions [2], [3], [4], [5]. Tumors are broadly classified either as benign or malignant, depending on their severity. The following three broad classes of tumors are usually known in the context of brain tumors, and they arise in the different parts of the brain: glioma, meningioma, and pituitary tumors[6], [7], [8].

Early detection and categorization of the tumors is highly and significantly important to patient survival and an immediate and early intervention [9], [10], [11]. This may result in a benign tumor developing into a malignant form due to a delay in treatment. Two key determinants of diagnostic precision are usually influential. The former is associated with the quality of the medical image, which can be negatively affected by the presence of specularity, saturation, artifacts, blur, low contrast, bubbles, or instrument and blood interference in this way, obstructing the proper detection [12], [13], [14]. The second is through human constraints, since radiologists can misclassify the tumors either due to fatigue, heavy workload, cognitive bias, or their diffuse levels of expertise.

This requires an automated diagnosis system that will assist human experts in their reliability in classifying brain tumors. This system should have been able to automatically draw deep features on medical images, helping to note the existence of subtle changes in texture and other repetitive patterns of texture. The system is to be trained and tested in relatively large and varied domain-specific datasets to be able to perform robustly. Nevertheless, individual medical records and medical facilities are not frequently eager to open their medical information to even research purposes because of the privacy issues [15].

Previously, machine learning (ML) models were actively used in the classification problem. Nevertheless, such methods had significant shortcomings [16], [17], [18], [19]. Not only did they need manual feature engineering, but they also could not give any deep feature representations, and the evaluation metrics they did generate were inadequate to use in the clinic. Therefore, studies shifted towards deep learning (DL) models, which overcome these weaknesses by extracting features in the form of hierarchy and automaticity[20], [21], [22], [23], [24]. However, the ML to DL transition presents two immediately important issues: First, the computational cost, and second, the training data volume requirement[25], [26]. The initial difficulty has mostly been reduced by the supply of modern multi-core processors and processors with graphics cards. The second difficulty is somewhat overcome by gathering bigger sets of information or using data augmentation.

Deep learning architecture (pytorch.org) offers a successful variety of models of different complexities (Appendix A(a)). As an example, DenseNet201 has around 20 million parameters, and the cost is 4.29 GFLOPs, so ViT-B/16 has a cost of 17.56 GFLOPs and has around 86 million parameters. Such variability indicates the need to choose a suitable model based on the task and the amount of available computational resources. Convolutional Neural Networks (CNNs) have been especially effective at local feature capturing, such as example local boundaries and sides, whereas transformer-based networks are effective at capturing long-range interactions and features in the global context.

Vaswani et al[27] proposed the first transformers to be used in tasks of natural language processing due to the efficiency of representations of long-range dependencies. To tune transformer architectures to computer vision, pictures were broken down into fixed-size patches (e.g., 16x16) as suggested by Dosovitskiy et al.[28], resulting in the design of Vision Transformers (ViTs). ViT models work especially well at capturing such global imagery features as overall color plus shape and texture [29], [30], [31].

Currently, CNN and ViT, when used in different architecture forms, utilize their disadvantages to leverage their advantages. Hybrid models are suggested to harness these complementary abilities to combine CNNs with ViTs to achieve better feature representation and classification accuracy[32], [33], [34]. The misclassifications, false positives (FP), and false negatives (FN) are not equally important in all areas of application that include medical, natural, and social [35], [36], [37]. False negatives are especially critical in the context of a critical medical diagnosis, because an undiagnosed case, like that of cancer, may result in ineffective treatment administration, and the patient will have reduced chances of survival. In other arenas, on the other hand, false positives may be more vital. An example of this is the email spam filtering, whereby a crucial message is listed as spam, hence causing missed communication.

Models, which have been trained in natural image databases, cannot be directly used in the area of medical imaging since the medical images have unique characteristics and properties [38]. Therefore, one must provide models with a large and representative amount of medical data. Nonetheless, privacy in the health sector can be a hindrance to large-scale data sets of sufficient examples. To overcome this issue, different methods like data augmentation and synthetic image generation are widely used to diversify the dataset and increase the generalization ability of the deep learning models [39], [40], [41], [42].

In the present investigation, an adequate set of data is put together that includes three popular types of brain tumors and a non-tumor group. The tumor types are classified in terms of their cellular origin as well as their severity. Gliomas are formed in the glial cells, which support and provide neurons, and are the most aggressive and severe types of brain tumors. Meningiomas form on the meninges, which are the three layers of tissue that sit between the skull and the brain, and will usually not be very aggressive, but may cause great complications. The pituitary gland is a tiny yet essential endocrine gland at the bottom of the brain, which forms tumors, causing pituitary tumors, and may lead to hormonal complications. The proposed study has the following significant contributions.

- Proposes two tumor-aware experimental paradigms: (i) a Boosted Feature Space (BFS) that independently optimizes DenseNet and Swin_t representations for high-sensitivity glioma detection, and (ii) a hierarchical DenseNet-Swin_t architecture with Deep Feature Extraction and Dual Residual connections (DFE+DR) specifically designed to suppress false negatives in meningioma and pituitary tumor classification.
- In the hybrid BFS setup, independently customized DenseNet and Swin_t models learn complementary local textures and long-range contextual dependencies; their dimension-

- aligned features are concatenated and boosted to enhance discriminative capacity for diffuse glioma patterns.
- In the DFE+DR setup, DenseNet functions as a stem CNN for structured deep feature extraction, followed by a customized Swin_t for hierarchical global reasoning, with dual residual connections stabilizing feature propagation and suppressing false negatives in meningioma and pituitary tumors. Finally, integrate a local–global feature of DFE+DR and BFS results that yield a unified correlated representation, improving robustness, class separability, and clinical reliability across heterogeneous tumor morphologies.
- The proposed EDSH framework demonstrates consistent performance gains and reduced false negatives on a large-scale MRI dataset (40,260 images across four tumor classes), confirming robustness and translational applicability, and compared with existing CNNs/ViTs and hybrid models. The dataset contains a large number of heterogeneous, stringent examples to allow strong training and generalization.

The remaining paper is structured as follows: Section 2 of the paper is a survey of the related work, Section 3 is the structure of the proposed framework, Section 4 is the implementation details, Section 5 is a report of findings (results) and extended discussion presented there, and the paper is concluded in Section 6.

## 2. Related Work

The researchers using CNN-based models have demonstrated significant accuracy in the classification of brain tumors. Indicatively, the model in [44] had an accuracy of 91.4% on publicly available data. Sajjad et al. [45] also increased the performance further to 94.58% using data augmentation and fine-tuning using a Softmax classifier. Similarly, to achieve 97.2% accuracy, Ahmet and Muhammad [46] scaled the ResNet-50 with alterations on the final layers. Sultan et al. [47] tested their CNN model on 3,064 MRI scans related to glioma, meningioma, and pituitary tumors, and obtained 96.13% accuracy.

In a different work, Ozyurt et al. [48] combined AlexNet feature extraction with an SVM classifier after using the NSEMFSE-based segmentation to achieve a detection rate of 95.62%. Sajja et al. [49] used VGG16 on the BRATS data to discriminate among malignant and benign tumors with an accuracy of 96.7%. Irrespective of these encouraging findings, CNNs have fundamental limitations driven by their local receptive fields, powerful inductive biases, and short-term ability to discern long-run or global associations. Therefore, they can have difficulties with differences in nature and contrast between MRI scans. Transformer-based architectures are recent developments in brain tumor classification studies to manage these shortcomings.

A vision transformer model has been trained from scratch on a multi-class classification of gliomas, meningiomas, and pituitary tumors using T1-weighted MRI scans (Figure 1: common modalities of MRI, Figure-2 MRI planes). The interpretability methods that were included in this model, Grad-CAM, LIME, and SHAP, had an accuracy of 91.61%, which was significantly better than the 83.37% of CNNs, but the scaling was also limited due to high data and computational requirements [50]. To reduce these limitations, the Hybrid model has been designed, which involves extracting local features from CNN modules and to learn and capturing global relationships with Transformer blocks.

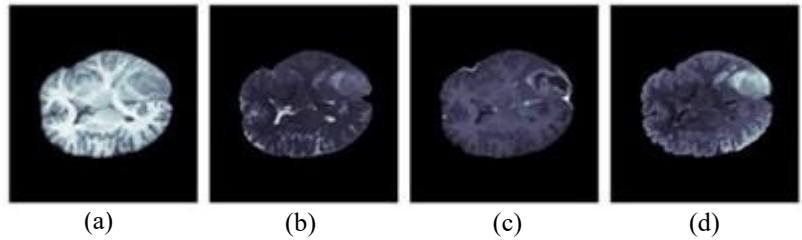
**Figure-1**: MRI images common Modalities (a) T1 (b) T2 (c) T1c (d) FLAIR

Aloraini et al.[34] offered a hybrid model, which is a combination of transformers and CNNs, and tested it on the BraTS2018 dataset, with an accuracy of 96.75%, but casts doubts on its applicability to heterogeneous groups. Bataineh et al.[51] come up with a hybrid model with the introduction of relative position encoding and residual structures to MLP layers, having Swin Transformer and ResNet50V2 to extract global and local features of the brain tumor MRI images. A benchmark dataset from Kaggle containing four classes is used, and a 96.80% accuracy result is obtained. Oztel et al.[52] ran an ensemble technique that comprised Vit and CNN models on the Sartaj dataset (Figure-2 MRI images planes). the highest accuracy was 85.03%, the ensemble comprised of MobileNet-V3, Vit(d), ResNeXt, DenseNet201 and Vit.

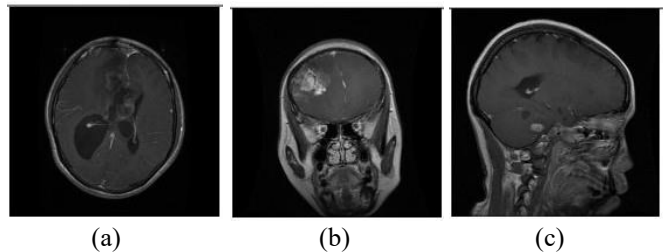
**Figure-2**: MRI images three Planes (a) Axial (b) Coronal (c) Sagittal

Based on the literature review, it is possible to note two significant weaknesses: a small amount of data and inadequacy in terms of evaluating measurements (Table 1). Various examples are required in a sufficient number to reduce the overfitting and enhance the generalization of the model. Sensitivity (recall) is more important in the medical field compared to most other measurement areas. To illustrate from literature, RT-PCR testing for COVID-19 was reported to have a low sensitivity level, and a high rate of false negatives, subsequently leading to the use of probing as a first anamorphic screening method, whereas CT imaging was utilized for final evaluation, as it had a high sensitivity rate [35], [36], [52]. The same is true with brain tumor diagnosis, where false negatives should be minimized to the highest degree.

Hence, the target research of interest is:
- Creating an enhanced dataset.
- Sensitivity is the most important measure of evaluation.

Table 1: Literature review of recent studies

| Author | Architecture | Dataset Size (MRI) | Tumor Classes | Accuracy/ Recall | Limitation |
|---|---|---|---|---|---|
| Togacar et al. (2020) [9] | CNN model BrainMRNet with attention module and hypercolumn | 253 MRI images | 2 (Tumor, Non-tumor) | BrainMRNet: -Accuracy 96.05% -Recall 96.0% GoogleNet: -Accuracy 96.05% -Recall 96.0% | -Only binary classification. -Limited dataset. -Low accuracy (as two classes). |
| Pokhrel et al. (2022) [53] | MobileNetV2, V3, VGG (16/19) | 3000 images | 2 (Tumor, Non-tumor) | MobileNetV2/(V3: -Accuracy : 94.17%/94.38%. VGG (16/19): -Accuracy: 93.7%, 92.57%. | -No data augmentation. -Binary classification. -No recall computed. |
| M. Aloraini et al. (2023) [34] | Hybrid Transformer-Enhanced CNN | 2018 images | High-Grade & Low-Grade Glioma | TECNN: -Accuracy: 96.75% -Recall: 97% | -Relatively small dataset |
| Hossain et al. (2024) [38] | InceptionV3, Vgg16 & 19, Xception, ResNet50, Inception-ResNetV2, IVX16, Swin, CCT and EANet. | 3264 MRI images | 4 (Glioma, Meningioma, Pituitary, Normal) | IVX16: -Accuracy: 96.94% - Max Recall: 79% Vgg16: -Accuracy: 95.11% | -Limited dataset. -Very low recall, 79%. |
| S. Hong et al. (2024) [43] | HF, CLAHE, and UM image enhancement techniques +, Modified VIT-B/16 | 7023 MRI images | 3 (Glioma, Meningioma, Pituitary) | VIT-B/16: -Accuracy: 91.36% | -Small dataset. -Healthy MRI images not included. -No recall metric computed. |
| Bataineh et al. (2024) [51] | Hybrid model: Swin Transformer and ResNet50V2 | 7023 MRI images | 4 (Glioma, Meningioma, Pituitary, Normal) | Hybrid model: -Accuracy 96.8% -Recall 97.0% Resnet50V2: -Accuracy 91.1% -Recall 91.0% | -Limited dataset. -Only one hybrid model is tested. |
| Shamshad et al. (2024) [54] | TL with VGG-16, ResNet-50, MobileNet using CNNs | 256 images | 2 (Tumor, Non-tumor) | ResNet50: -Accuracy: 96.0% -Recall: 90% MobileNet: -Accuracy: 87% -Recall: 79% | -Small dataset -Binary classification -Poor generalization. |
| Oztel et al. (2025) [52] | Bagging method using MobileNetv3, vision transformer, ResNeXt, and DenseNet-201. | 3264 MRI images | 4 (Glioma, Meningioma, Pituitary, Normal) | Bagging all: -Accuracy 85.03%. Only two model bagging: -Accuracy: 83.25% | -Recall value not computed. -Limited dataset |
| Mzoughi et al. (2025) [55] | Vision Transformer (ViT) and CNN Comparative Study. | 3064 MRI images | 3 (Glioma, Meningioma, Pituitary) | ViT: -Accuracy: 91.61% -Recall: 98% CNN: -Accuracy: 83.37% -Recall: 83.5% | -Normal MRI not included. -Low-resolution images. -Inconsistent results. |

## 3. Proposed Methodology

This work proposed a specifically designed efficient DenseNet-Swin Transformer Hybrid (EDSH) framework that analyzes images of brain tumour MRI by analyzing fine-scale texture features and long-range contextual clues. There are two tumour-aware configurations developed to address the diagnostic complexity of classes. The proposed work first combines a Boosted Feature Space (BFS), which better learns the irregularity of tumor shape, tumor heterogeneous texture, and poorly defined mass of tumor (thus better classifying glioma).

The latter structure utilizes a hierarchical DenseNet Swin_t, resulting in a Deep Feature Extraction with Dual Residual connections (DFE+DR), which better learns well-defined and homogenous tumors (meningioma and pituitary). During prediction, both the BFS and DFE+DR predictions are task-specific at the output stage, which are integrated systematically by the SI module to maximize class separability and clinical robustness in a wide spectrum of tumour morphology. The proposed hybrid model architecture is shown in Figure-3 while a comparison with existing CNNs, Vits, and Hybrid classification models is done in the following pages, and according to that block design is shown in Figure-4.

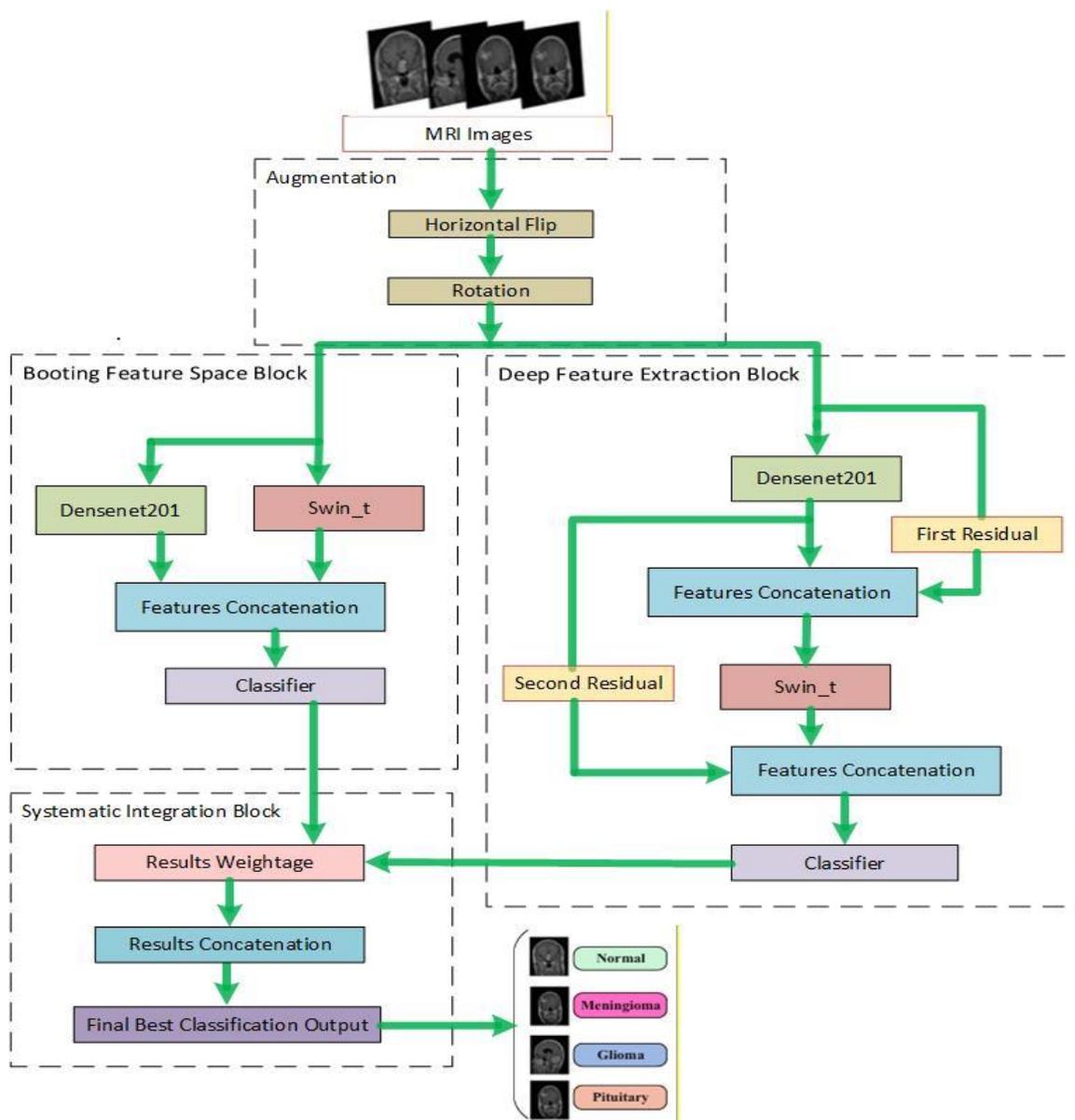

**Figure-3**: Proposed EDSH framework architecture

## 3.1 Preprocessing

Data augmentation is often utilized to increase the generalization and reduce the problem of model overfitting [11], [38], [60]. Only two augmentation methods are used on the original data in the current study: horizontal flipping and random rotation (Table 2). As the dataset already has enough examples of the domain, these two augmentation strategies are sufficient to increase variability without adversely replicating important anatomical features. The probability of the horizontal flip is 0.5, which means that every image has an equal chance of being flipped, whereas the random rotation is 10 degrees, so that images could be rotated randomly between the range of ±10 degrees. The MRI images are collected from different available sources, thus their dimensions vary. So, to make them suitable for model input, each image is cropped to a standard 224 x 224 size.

**Table 2**: Data Augmentation applied

| Parameter | Value |
|---|---|
| Horizontal Flip | ±0.5 |
| Rotation | ±10 |

## 3.2 Boosting Feature Space

Convolutional Neural Networks (CNNs), considered to be black-box models, are very useful in capturing local features in images but overlook long-range associations [56]. In order to mitigate feature redundancy and enhance the information flow, it was proposed by Huang et al.[57] 2018 in the DenseNet architecture, where a cascaded connectivity structure reuses the features (Eq (1)) that are extracted at a certain layer in the next layer (Figure-4). Zahoor et al.[2] compared various classification models and discovered that DenseNet201 had a competitive result, and it is the top-ranked as the runner-up on a benchmark dataset with brain tumor classification.

In our proposed EDSH framework, therefore, we utilize DenseNet201 as the CNN component, as it has moderate computational complexity, but good feature extraction (Appendix Aa(0)). The Densenet201 learns features like local texture, homogeneous or heterogeneous, margins, smooth or rough, shape, irregular or regular, and for spatial features, learns local sub-regions. As for the current study on brain tumors, the following local features are learned: tumor margin sharpness, internal texture, peritumoral edema patterns, edges, enhancing core/non-enhancing, tumor shape, mass effect, tumor sphericity, and surface roughness.

$$y = Conv.\ Relu\ .BN(x_0, x_1, x_2 \dots \dots \dots) \qquad (1)$$

Transformers' unprecedented success in language translation and other problems with long-range dependencies[58]. In the same success, images are modeled as a sequence of 16x16 patches, and developed Vision Transformers (ViT). ViTs showed high-quality results in a range of image classification benchmarks, but their disadvantage is that they are computationally costly [56]. To overcome this weakness, Z. Liu et al.[59] proposed in 2021, a hierarchical ViT model named the Swin Transformer, a shifted window scheme to achieve a balanced efficiency-accuracy model. Therefore, for computational efficiency, the Swin Transformer (Swin_t) model is used as a transformer component of the proposed EDSH framework in the present research. Secondly, it compares in level of complexity to the DenseNet201 CNN model employed in the proposed EDSH framework (Appendix A(a)).

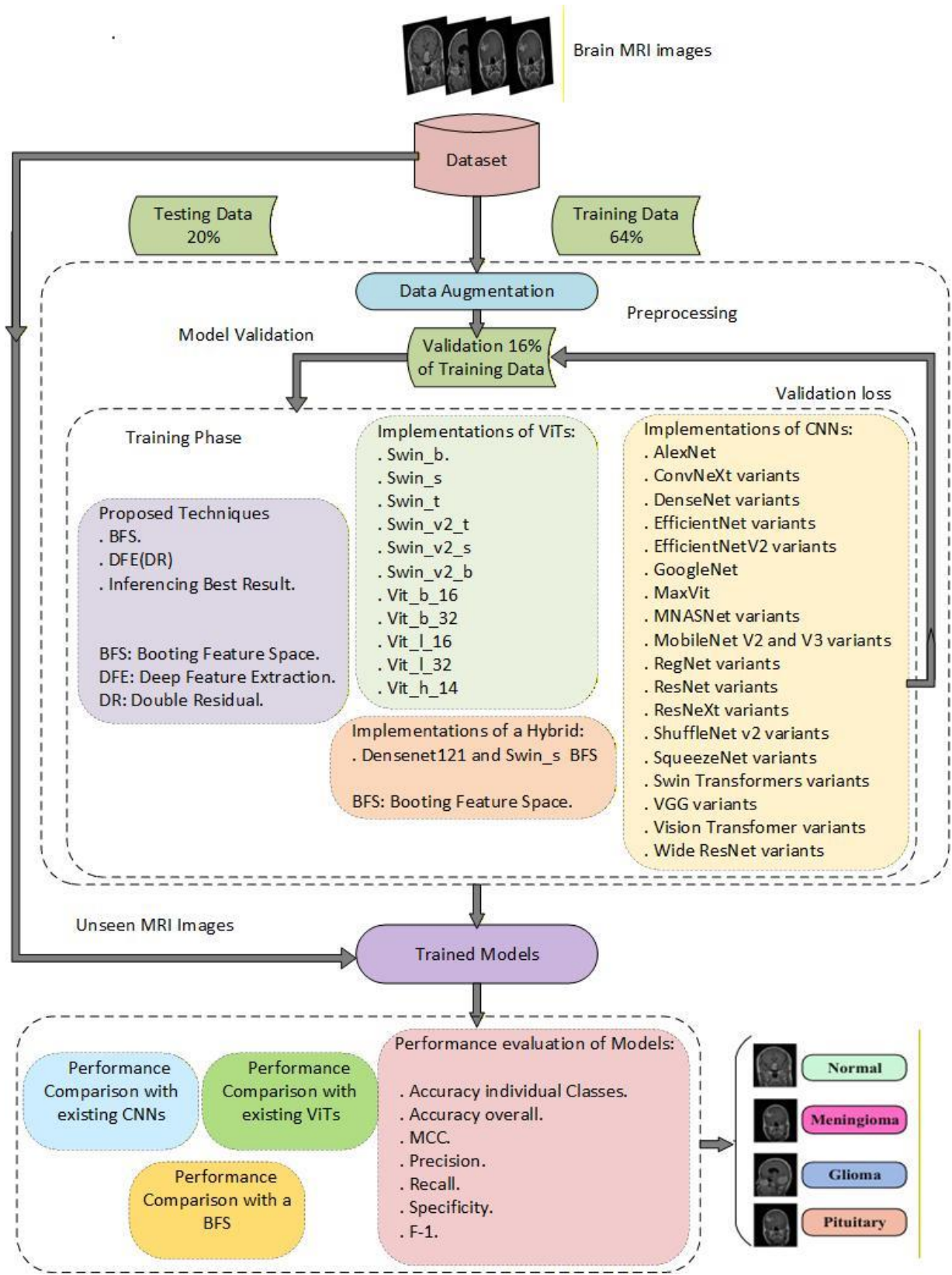

**Figure-4**: Block design of Proposed EDSH framework

The Swin Transformer model Swin_t captures the contextual features like global context, spatial relationship, multi-scale features, and sequence alignments. From the perspective of brain tumor MRI images following global features are learned; relationsip between tumor core and distant edema, cross-hemispheric effect, anatomical neighborhood awareness, symmetry or asymmetry between brain hemispheres, most aggressive tumor sub-regions, peri-tumoral infiltrative zones, and multi-focal tumor connectivity.

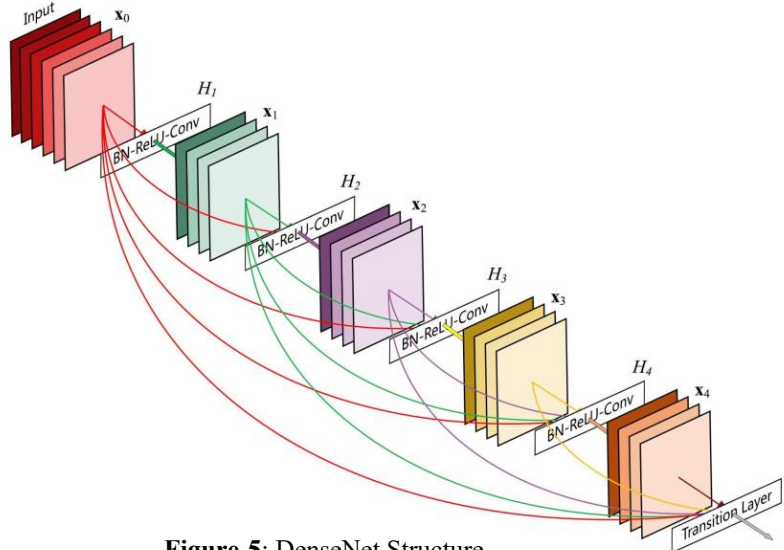

**Figure-5**: DenseNet Structure

The initial experimental design proposes a Boosted Feature Space (BFS) framework (Figure-6), whereby disjointly customised DenseNet and Swin Transformer (SwinT) branches are formulated to acquire dissimilar local and global depictions of MRI scans of the brain. The DenseNet branch has the responsibility to capture fine-scale spatial textures and edges of tumor boundaries through dense propagation of residual features, and the Swin Transformer branch can capture long-range contextual associations and global tumor morphology through window-based self-attention. An input MRI picture X is put into DenseNet to obtain a local feature embedding $F_D \in R^{N \times d}$ and Swin_t to obtain a global feature embedding $F_S \in R^{N \times d}$. To ensure compatibility, linear projections are used to ensure that both feature spaces are dimension-aligned, defined in Eq-2.

$$\overline{F_D} = F_D W_D, \overline{F_S} = F_S W_S \qquad (2)$$

$$F_{BFS} = \alpha \overline{F_D} \oplus \beta \overline{F_S} \qquad (3)$$

The aligned features are then combined through concatenation and weighted boosting, where α and β are parameters to be learned to manage the effect of local and global information, Eq. 3. This design of boosting by representation makes the diffuse and infiltrative patterns of glioma more sensitive to discrimination, which in most cases have weak edges and non-uniform textures. The BFS set-up, which delivers the end-to-end joint optimisation of both branches, effectively boosts the set-of features that are diagnostically significant and down-regulates the number of redundant responses, thereby enhancing class separability and robust glioma detection ability.

The glioma tumor MRI images have the features of high heterogeneity due to a mixture of viable tumor, necrosis, edema, and hemorrhage, a complex disordered texture which results in a high entropy and contrast, the shape and morphology of the tumor have fuzzy margins and poorly defined edges, and also low sphericity. The BFS CNN part learns the texture changes of peritumoral edema extent, local texture variations of enhancement heterogeneity, local ventricle compression of mess effect severity, and edge sharpness for the tumor white matter interface. The Swin transformer part extracts the global relationship: global spatial spread patterns for peritumoral edema extent, regional enhancement patterns for enhancement heterogeneity, global midline shift for mass effect severity, and infiltration into adjacent tracts for tumor white matter interface. When these complementary features are concatinated then resulted in a boosting of glioma classification.

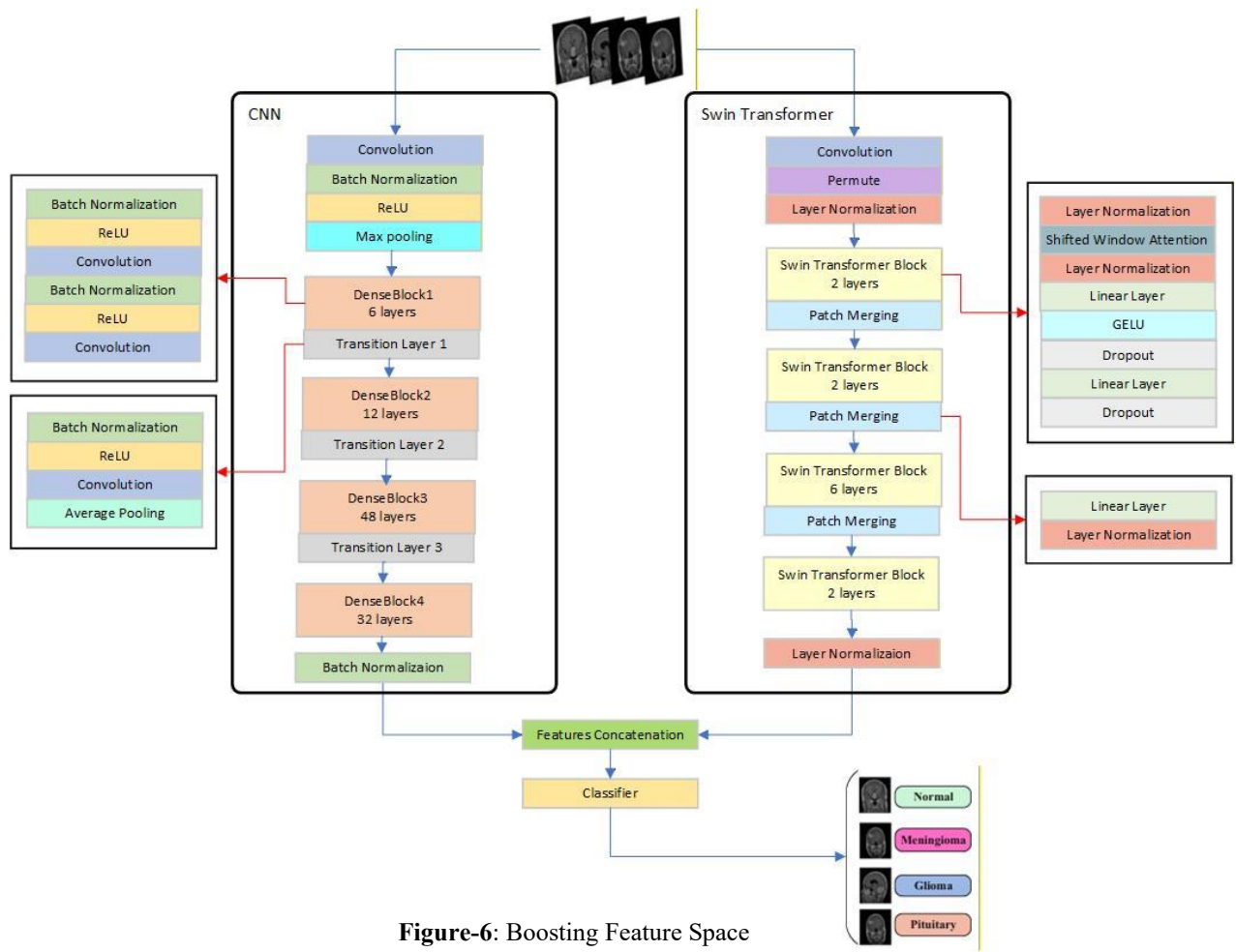

**Figure-6**: Boosting Feature Space

## 3.3 Deep Feature Extraction with Dual Residual Connections

The second experimental design is based on a hierarchical, DenseNet-Swin Transformer (Swin_t) architecture, including Deep Feature Extraction and Dual Residual connections (DFE+DR), to simultaneously exploit the structured local features and the global contextual representation in the analysis of brain MRI (Figure-7). Through careful customization of DenseNet at the input stage, spatial specificities of MRI images that necessitate the accurate extraction of fine-grained texture and intensity variations, which are essential in the accurate characterization of the tumor, have been made possible. This is achieved through dense residual connectivity that allows excellent feature recycles across consecutive layers, yet with detailed spatial information, at the same time reducing vanishing-gradient effects that are typical of training deep networks.

Simultaneously, it has been demonstrated that the Swin Transformer (Swin_t) has been purposely designed by using a task-oriented patch embedding to provide a smooth compatibility with the resolutions and tumor size that medical imaging may require. Moreover, shifted-window self-attention is incorporated, enabling the model to skillfully represent hierarchical worldwide dependencies through cross-window information interaction, and achieve computational scalability at the same time as inference of global tumor morphology is enhanced. Given an input image $X \in R^{H \times W \times C}$ a personalized DenseNet stem $\Psi_D(.)$, conditioned on an input image, can be seen as a convolutional feature extractor, which learns both low-level and mid-level spatial patterns and gives local feature maps Eq-4.

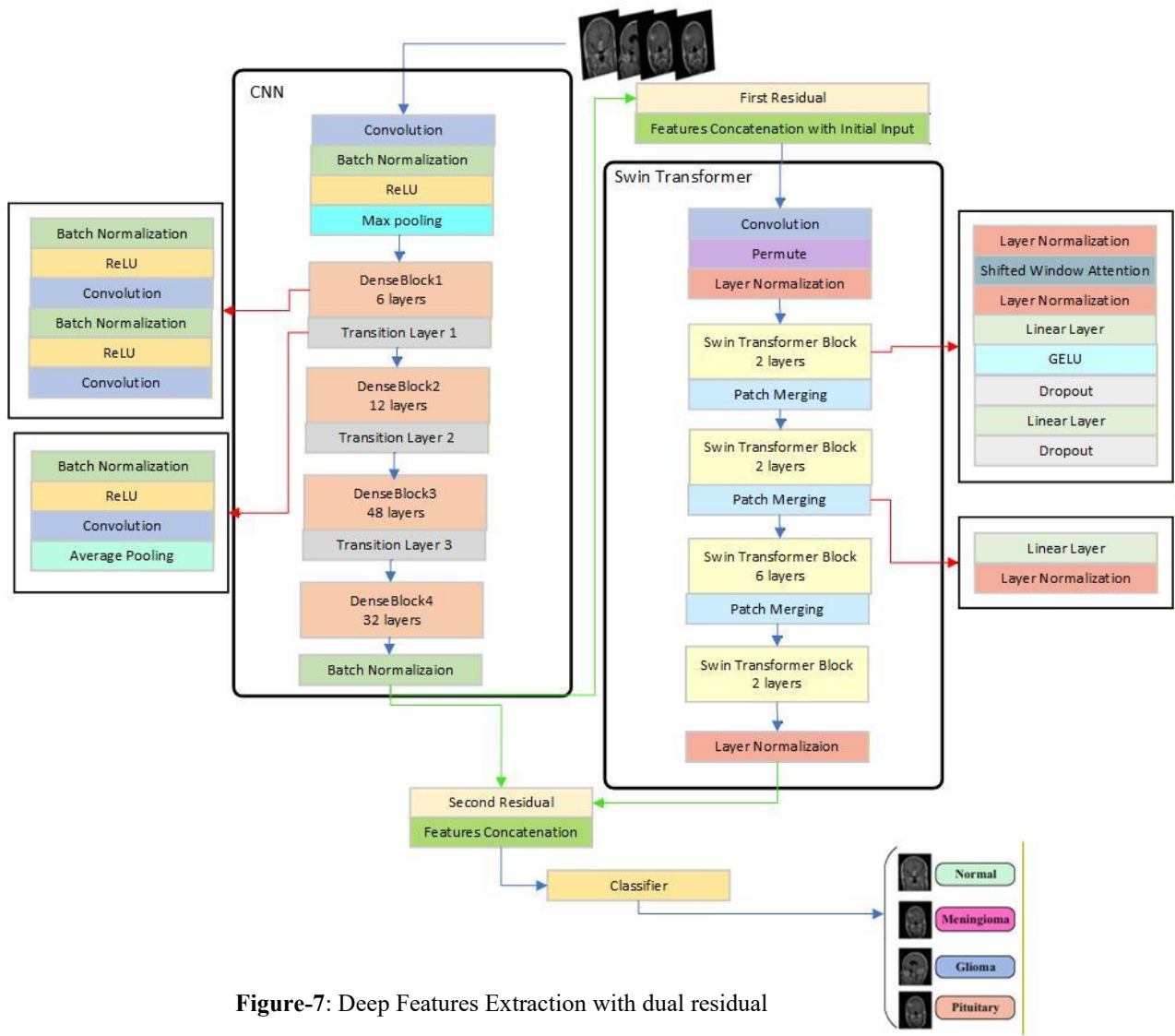

**Figure-7**: Deep Features Extraction with dual residual

$$F_D = \Psi_D(X) \in R^{H \times W \times C} \qquad (4)$$

To maintain original spatial cues and the propagation of gradients, a first residual connection is created through the concatenation of the output of the DenseNet and the raw input features of the images, where $\acute{X}$ represents the dimension-matched representation of the input. $R_1$ is then kept as the residual feature Eq-5 and sent to the Swin_t encoder, which implements window-based and shifted-window self-attention to learn long-range features and global tumour morphology Eq-6. To support deeper feature reuse and hierarchical fusion, a second residual connection is to be created by adding the output of Swin_t with an in-between feature representation of DenseNet. This rich residual feature space $R_2$ captures fine-grained texture details Eq-7 and worldwide structural context, minimizes the loss of important information, and, to some extent overcome vanishing gradient issue (vanishing gradient issue because of the proposed setup depth), and it is provided to the classifier to yield the final classification Eq-8.

$$R_1 = F_D \oplus \acute{X} \qquad (5)$$
$$F_S = \Psi_S(R_1) \in R^{N \times d} \qquad (6)$$
$$R_2 = F_S \oplus F_D \qquad (7)$$
$$\hat{y} = Softmax(R_2 W_C + b_c) \qquad (8)$$

By integrating both convolutional and transformer-based representations with hierarchical integration through two residual connections, the DFE+DR framework improves discriminative learning, in addition to stabilising optimisation. This design will be especially useful in the rejection of false negatives in the classification of meningioma and pituitary tumours. The MRI images of meningioma and pituitary have following features; meningioma have high homogeneity while pituitary vary from homogeneous to hetrogenous, meningioma have simple ordered texture with low contrast and uniform regions while pituitary have variable texture but more ordered than glioma, meningioma have smooth and well-circumscribed like spherical or oval with sharp margins while pituitary have 'snowman or 'figure 8' like shape or constrained by location.

The meningioma and pituitary tumor both occur outside the brain. As discussed in the features, both meningioma and pituitary are alike in some features, but are very different from glioma. In the DFE+DR hierarchical structure, CNN serves as a 'local feature dictionary creator' and swin transformer serves as a 'feature relationship matcher', swin transformer never processes raw pixels but processes semantically meaningful CNN features therfore if CNN miscalssifies a local region, swin transformer re-contextualizes it. The residual connections served these purposes; the first residual connection prevents Densenet from over-transforming, second residual ensures Swin_T would not lose CNN local insights.

## 3.4 Systematic Integration Block

At the target level, the task-specific output of the Boosted Feature Space (BFS) and the Deep Feature Extraction with Dual Residual connections (DFE+DR) structures are consolidated systematically with the help of a special Systematic Integration (SI) block (Figure-8), which results in higher classification separability and patient-specific reliability to all heterogeneous brain-tumor morphologies. Let $P_{BFS}$ represent the probability vector of the result of the BFS branch that is optimised in diffuse and infiltrative glioma characterisation due to its enhanced combination of local and global representations. Likewise, let $P_{DFE}$ represent the result of the DFE+DR branch, which has an improved sensitivity of small and well-defined tumors like meningioma and pituitary adenomas. In order to harness both frameworks' strengths to complement each other, the SI block does a weighted, target-level fusion which is defined in Eq. 9, where $\lambda \in [0,1]$ is a learnable or class-adaptive integration coefficient. The final class prediction is in Eq. 10.

$$P_{SI} = \lambda\, P_{BFS} + (1 - \lambda)P_{DFE} \qquad (9)$$
$$\hat{y}_{SI} = \arg\max(P_{SI}) \qquad (10)$$

This integration approach is a selective amplification method to optimally accurate predictions, by each branch, allowing strong discrimination between tumour subtypes with varied morphological patterns. Through a combination of the enhanced sensitivity of BFS in the classification of glioma with the high specificity of DFE+DR in terms of meningioma and pituitary tumours, the SI block will provide better diagnostic consistency and minimise the number of class-dependent misclassifications in a unified decision support system.

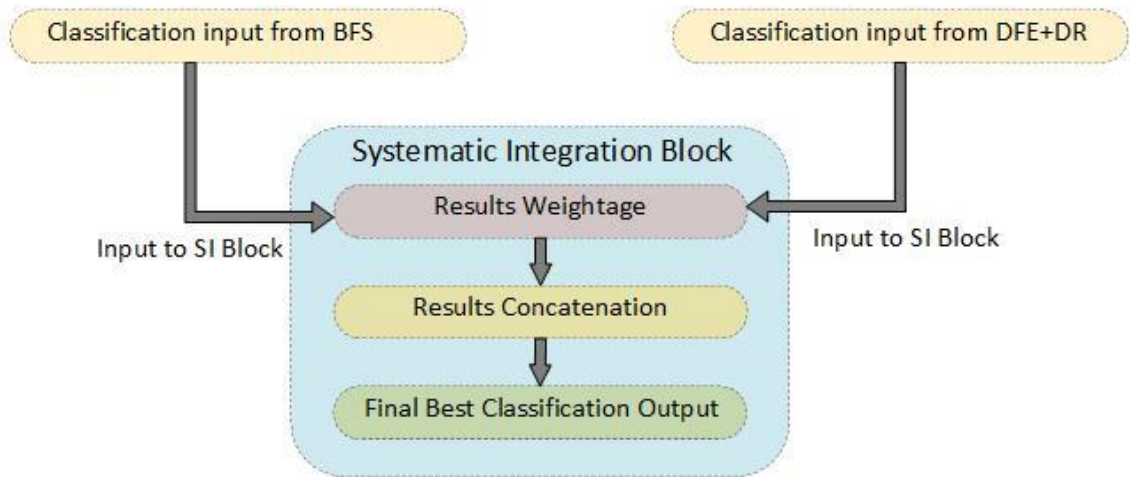

**Figure-8**: Systematic Integration Block

## 4. Implementation Details

### 4.1 Dataset:

Open-source benchmark repositories such as Kaggle and Figshare were used to compile the dataset of this research, and all the links to the source are in Table-3. The number of the gathered MRI images amounted to 40,260, which included four asthmatic classes, namely, three classes of tumor (glioma, meningioma, and pituitary) and one healthy (no-tumor) class. The data was split initially among the training (64%), validation (16%), and testing (20%) using the most popular method.

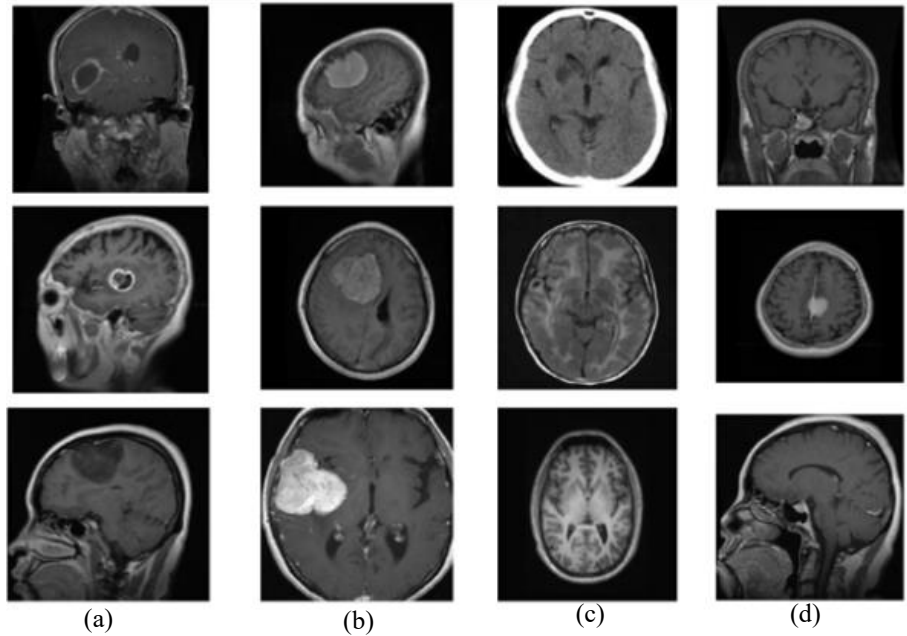

**Figure-9**: Some examples from dataset, (a) glioma (b) meningioma (c) no tumor (d) pituitary

**Table 3**: Dataset Sources

| Sr. No | Dataset Sources: | Modalities Description | Tumor Classes | No. of Samples |
|---|---|---|---|---|
| 1 | https://www.kaggle.com/datasets/thomasdubail/brain-tumors-256x256 | Serial "2" has T1, T1 with contrast, and T2. Serial "9" has multimodalities, as in literature "Ghosh et al.2023"[10] Sartaj T1, T2, and FLAIR. All other sources, modalities, and details are unavailable. | Glioma | 10,357 |
| 2 | https://www.kaggle.com/datasets/fernando2rad/brain-tumor-mri-images-17-classes | | Meningioma | 10,638 |
| 3 | https://www.kaggle.com/datasets/alaminbhuyan/mri-image-data | | no-tumor | 7937 |
| 4 | https://www.kaggle.com/datasets/mohammedhamdy98/brain-tumor-mri-classification | | Pituitary | 11,328 |
| 5 | https://www.kaggle.com/datasets/aqibrehmanpirzada/brain-tumor-classification-dataset | | | |
| 6 | https://www.kaggle.com/datasets/mohammadhossein77/brain-tumors-dataset | | | |
| 7 | https://figshare.com/articles/dataset/brain_tumor_dataset/1512427/5 | | | |
| 8 | https://www.kaggle.com/datasets/mohamedmetwalysherif/braintumordataset | | | |
| 9 | https://www.kaggle.com/datasets/sartajbhuvaji/ | | Total | 40.260 |

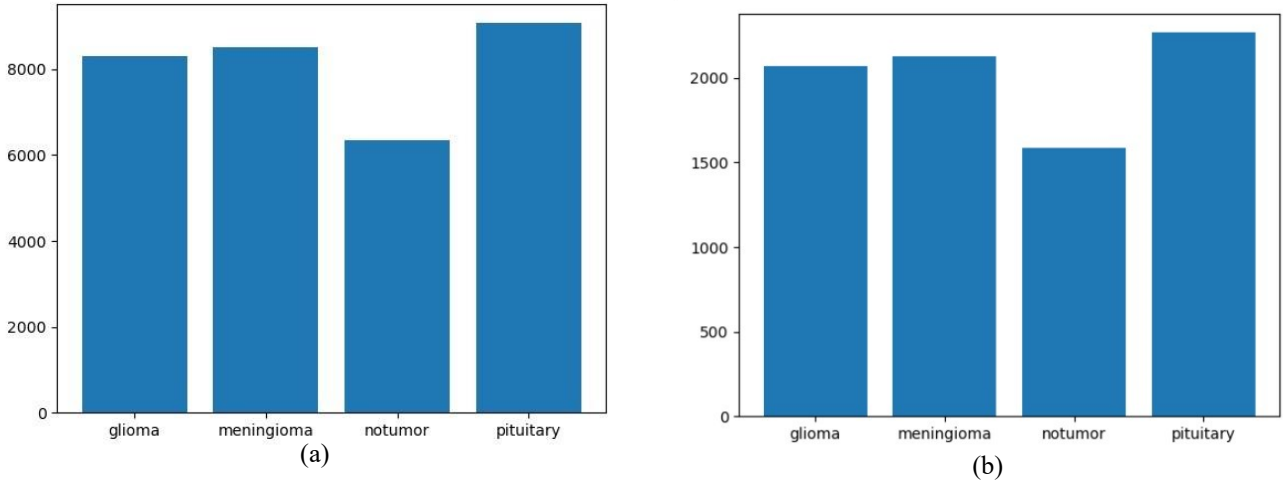

**Figure-10**: (a) Training (64%) + Validation (16%) and (b) Test (20%) dataset bar-graph.

**Table 4**: Train, Validation, and Test Distribution of the Dataset

| No. of Classes | Training | Validation | Testing | Total |
|---|---|---|---|---|
| **Glioma** | 6,629 | 1,657 | 2,071 | 10,357 |
| **Meningioma** | 6,808 | 1,702 | 2,128 | 10,638 |
| **Normal** | 5,080 | 1,270 | 1,587 | 7,937 |
| **Pituitary** | 7,250 | 1,812 | 2,266 | 11,328 |
| Total | 25,767 | 6,441 | 8,052 | 40,260 |

proportion[30]. The split of MRI images in the four categories is in Table 4, and the corresponding bar graphs are in Figure-10. In terms of the balance of the classes, it is possible to see that there is no significant imbalance in the dataset, as from the dataset, the meningioma is up to 3% bigger than the glioma, and the pituitary is up to 9% bigger.

The dataset was further investigated in order to detect possible redundancies. It was noted that the data sets of Thomas and Mohammad Hossein (Sr. Nos. 1 and 6 in Table 3) are all from the same dataset, Sartaj. But both authors used different image enhancement and image augmentation methods to obtain modified versions. Thomas processed: image normalization, resizing with aspect ratio preservation, and redundant data removal to improve the quality of the image. The author Mohammad Hossein, on the other hand, used several forms of data augmentation and enhancement, which were salt-and-pepper noise addition, histogram equalization, rotation, adjustment of brightness, and horizontal and vertical flipping to enhance the diversity of data.

The Multimodal MRI examples (T1, T2, and FLAIR) are also present in the dataset, which is specifically noted in the datasets by Fernando Feltrin (Sr. No. 2) and Sartaj Bhuvaji (Sr. No. 9) in Table 3. Moreover, there are all three standard anatomical orientations, including axial, coronal, and sagittal viewpoints, included in the dataset. The images collected are not in the same resolution. So, each of the images was cropped and resized to 224 x 224 pixels so that it can be compatible with the model input format. Two data augmentation methods have been used, namely horizontal flipping and random rotation. This increased the number of images in the dataset to a total of 120,780 images (40,260 original images + 80,520 augmented images).

## 4.2 Experimental Setup and Implementation:

The proposed EDSH framework and all other models employed in the given research were trained in the following settings of hyperparameters: Stochastic Gradient Descent (SGD) as the optimizer, learning rate set to 0.001, and cross-entropy loss that helps to eliminate further effects of class imbalance. The batch size was set to 16, and each model took 50 epochs to train. All tests were performed on a Dell OptiPlex 7070 workstation with the Intel Core i7 (9th generation) processor and 64 GB RAM. The open source deep learning classification framework selected was PyTorch. The data was separated into training (64%), testing (20%), and validation (16%) sets.

The selection of hyperparameters was done by evaluating a base CNN model of our proposed EDSH (Densenet201) on a single change in hyperparameter Table 5. The first row is the results on hyperparameters that are selected for the current study, and then, one by one, in the following rows, the hyperparameters are changed. In the second, third, and fourth rows, batch size, optimizer, and learning rate are changed, respectively, while other hyperparameters remain the same. It can be concluded from that the hyperparameters optimizer: SGD, batch size: 16, and learning rate:0.001

Table 5: Selection of Hyperparameters

| Densenet201 (SGD, Batch Size=16, Learning rate = 0.001) | For all classes | | | | | | Individual Class accuracies | | | |
|---|---|---|---|---|---|---|---|---|---|---|
| | Accuracy | MCC | Precision | Recall | Specificity | F-1 | Glioma | meningioma | No-tumor | Pituitary |
| Unchanged | **97.33** | **96.33** | **97.33** | **97.33** | **97.93** | **97.33** | **96.33** | **96.83** | **98.23** | **97.83** |
| Batch Size = 32 | 80.33 | 72.33 | 79.33 | 80.33 | 91.33 | 80.33 | 54.33 | 80.33 | 96.33 | 89.33 |
| Optimizer = Adam | 78.33 | 70.33 | 78.33 | 78.33 | 90.33 | 78.33 | 60.33 | 63.33 | 97.33 | 93.33 |
| Learning rate = 0.01 | 96.33 | 95.33 | 96.33 | 96.33 | 97.63 | 96.33 | 92.33 | 96.53 | 98.13 | 97.93 |

give the best result, showing in bold in the first row.

The measures of evaluation in the study are Accuracy (individual and overall), Matthews Correlation Coefficient (MCC), Precision, Recall, Specificity, as well as F1-score. As it was mentioned above, the comparative significance of the evaluation metrics is based on the field of application[61], [62]. Recall is the most important metric in brain tumor classification since it is directly connected with the reduction of false negatives (FN). False negatives, in which a patient with a tumor is wrongly diagnosed as tumor-free, may have a serious clinical impact, as it might slow down the treatment process and permit a benign tumor to develop into a malignant one. Conversely, a false positive (FP) or the concept that a healthy person has been diagnosed with a tumor is a lesser harm, considering that it is usually considered for further diagnostic procedures where correct classification is attained.

$$Accuracy = \frac{TP+TN}{TP+TN+FP+FN} \quad (11)$$

$$Recall = \frac{TP}{TP+FN} \quad (12)$$

$$Precision = \frac{TP}{TP+FP} \qquad (13)$$

$$Specificity = \frac{TP}{TP+FP} \qquad (14)$$

$$F1 = 2 \times \frac{Precision \times Recall}{Precision + Recall} \qquad (15)$$

$$MCC = \frac{(TP.TN)-(FP.FN)}{\sqrt{(TP+FP)(TP+FN)(TN+FP)(TN+FN)}} \qquad (16)$$

## 5    Results and Discussion:

The different varieties of classification models present in the PyTorch framework were trained, validated, and evaluated on all the collected data with a standard set of hyperparameter configurations to avoid unfair comparisons. The results of the performance of the selected models are illustrated in Table-6. The proposed EDSH framework achieved the highest overall accuracy and recall when compared to CNNs, Vits, and some hybrid models. This result indicates the effectiveness of the EDSH framework in reducing false negatives since recall is the most important measure of tumor classification. The impact of using a large and diverse dataset is evident in the models with varying network depths.

### 5.1    Large Dataset Effect:

The varying depth models are selected from existing CNNs and Vits, from which the Vit_l_32 has the highest complexity, i.e., 15.38 GFLOPS and around 300 million parameters, and the Squeezenet1_1 has the lowest complexity, i.e., 0.35 GFLOPS and around 1.2 million parameters. The Vit_l_32 has the second-highest results in accuracy and in recall, Table 6, thus concluding that the depth has a direct contribution to the results because of the larger dataset. Also, from the results, it can be concluded that as the depth of the model increases, there is also a gradient explosion issue arises.

Two hybrid models and one ensemble model were also trained on the collected dataset for fair comparison, but some issues can be observed that hinder the good results from being achieved in these models. The computational requirements are increased because of the increased depth of models hybridized or ensembled as compared to the proposed EDSH framework. There is a gradient explosion issue due to the depth. The varying complexity models are hybridized or ensembled; hence, the output could not be concatenated well as in the proposed EDSH framework.  Therefore, all the models' hybrid/ensemble fail to achieve the best results, with some results even worse than individual CNN / Vit models.

### 5.2    Training and Loss Graphs

All the CNNs and Vits models under observation, train and validation accuracy curves are shown in Figure-11 (a) and (b) show the different CNNs and Vits training curves with the proposed EDSH framework. Our proposed EDSH framework train and validation accuracy curve does not show any overfitting or underfitting, and training runs very smoothly over the collected dataset; any kind of fluctuations are not observed in training over the complete path. Mnasnet1_0 and Googlenet show rigorous fluctuations because of their low complexity as compared to a large dataset, i.e., low complexity hinders the desired feature extraction. The proposed EDSH framework achieves its 80% of maximum peak on the 4th or 5th epoch, while other CNNs or Vits achieve their 80% on the 20th or 30th epoch.

The train and validation loss curves are shown in Figure-11 (c) and (d). The proposed EDSH framework shows the same best performance as in the accuracy curves. There is no fluctuation observed for the proposed EDSH framework. Also, the proposed EDSH loss curve, from 80 to 90%, becomes flattened on the 4th or 5th epoch. The other CNNs or Vits models have fluctuations in their loss curves. Some models' loss curves become 80% smoother at 20 or 25 epochs, while others remain unsteady. The proposed EDSH framework's accuracy and losses of training and validation become minimum around 42 epochs.

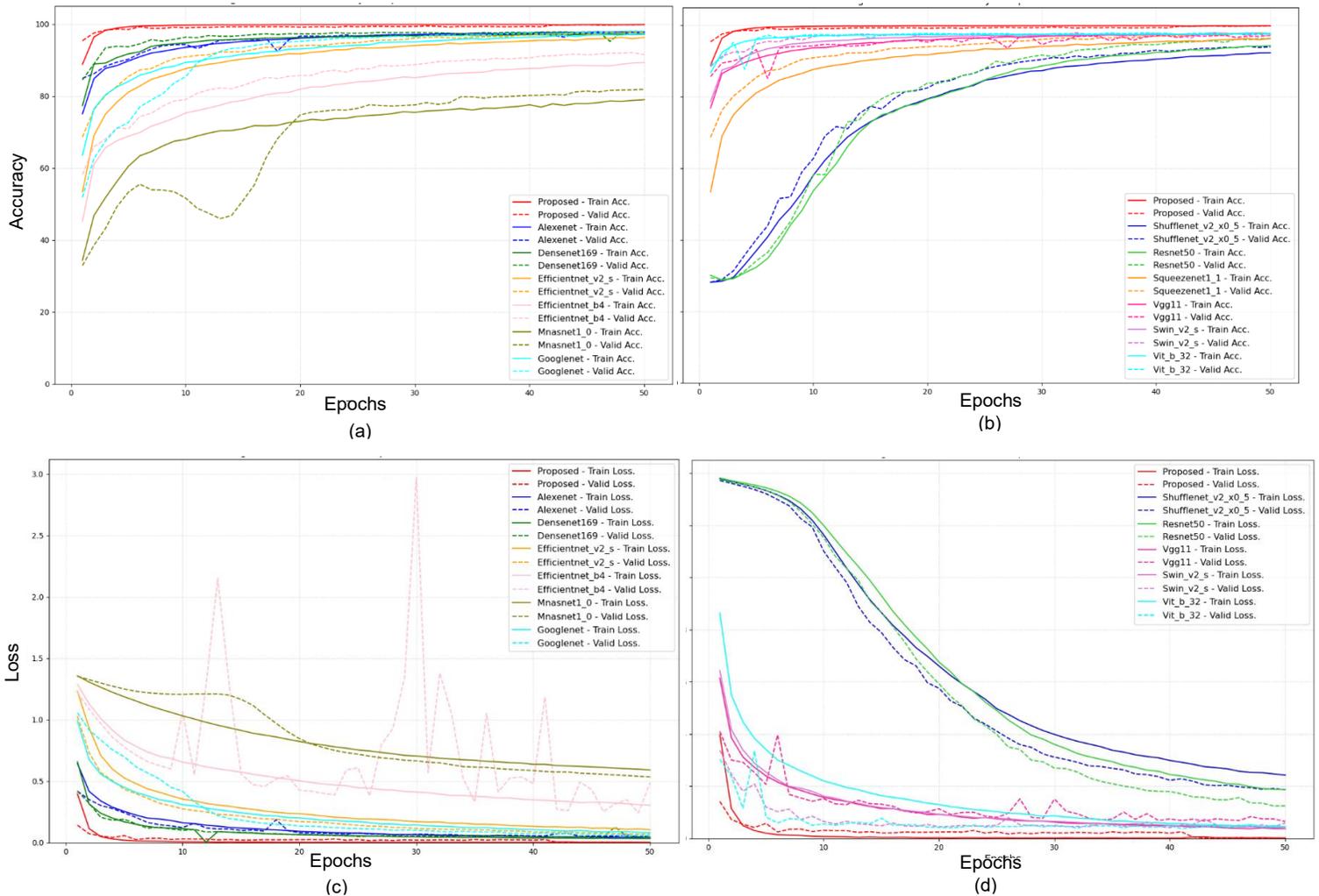

**Figure-11**: Comparision of Train and Valid, Accuracy and Loss of existing CNNs and Vits models with Proposed Hybrid Model.

### 5.3    Ablation Study

After individual testing of every component of the proposed EDSH framework, the overall model performance could not be repeated as in Table-6 and confirmed the ideal complementary entity of the proposed EDSH framework. The hybrid model, which was set to boost feature space (BFS) without deep feature extraction and residual connections (BFS-DFE-R), showed second best performance in results because complementary feature extraction adds performance to the feature space. The Deep Feature Extraction model without skip connections, without boosting feature space (DFE-R-BFS), yielded worse test performance as compared to boosting feature space because of complexity, and without residual connections. The introduction of a single residual connection into the DFE architecture without BFS (DFE+SR-BFS) did not add any performance gain as compared to BFS, and even performance degraded without a residual

By using the identical Deep Feature Extraction (DFE) architecture with two residual connections without BFS (DFF+DR-BFS), significant gains were realized in overall tests and in the individual class accuracies, the highest results achieved as compared to others already discussed. In the end, the proposed EDSH framework, by adding a boosting feature space model and deep feature extraction with a double residual model (BFS+DFE+DR) the highest results in accuracy and recall, as well as individual class accuracies.

## 5.4 Confusion Matrix

The confusion matrix of some selected CNNs, Vits, Hybrid, and ablation study models is compared with the proposed EDSH framework in Figure-12. The overall misclassifications for the proposed EDSH remain minimal compared to others. The vgg13, resnet50, vit_b_32, swin_v2_s, dense&swin1, dense&vit, BFS, DFE-R, and Proposed EDSH framework have FP; 124, 152, 95, 109, 118, 111, 64, 56, 51, and FN; 143, 182, 162, 266, 166, 155, 181, 103, 80, respectively. The proposed EDSH framework has a minimum of FP and FN, 51 and 80, respectively. The proposed EDSH is 10% less in FP from the minimum computed by other models (DFE-R: 56) and 30% less in FN from the minimum computed by other models (DFE-R: 103).

## 5.5 PR / ROC Curves

The proposed EDSH framework has been thoroughly analyzed and compared with existing CNNs/ViTs, and other representative hybrid model architectures using the Receiver Operating Characteristic (ROC) curve, Precision-Recall (PR) curve, and a model-wise gain analysis, as shown in Figure-13. The proposed EDSH framework has shown the highest area under the curve (AUC). In the same manner, the PR curves have a consistent trend; the proposed EDSH framework obtains finer accuracy at a greater range of recalls, which speaks to its solid operational performance in the situation of class-unbalanced conditions.

The gain analysis further confirms that our proposed EDSH framework provides the largest performance increment among all of the models under evaluation and therefore reflects its skillful features exploitation and the reliability of the decision-making. Several current hybrid structures, in turn, record relatively poorer results that can be explained by the greater complexity of architectural designs and training instability that can lead to gradient explosions and thus non-optimal convergence.

## 5.6 Feature Discriminative Analysis

The models listed in the ablation study, along with existing CNN, ViT, and hybrid models, have been visually compared with the proposed EDSH framework for feature-discriminative analysis, as shown in Figure 14. To be fair, all visualisations were created utilizing the same group of features extracted. The results indicate that the baseline models, in particular, ResNet50 and Vit-B/32, have significantly lower discriminative performance. Specifically, with ResNet50, the feature representations of each distinct class have a substantial overlap, indicating a lack of separability between classes and a poor representation of the discriminative information. Similarly, Vit-B/32 displays poor clustering as compared to more advanced hybrid formulations. On the other hand, the proposed EDSH framework gives very close centred clusters belonging to each tumor category with minimal overlapping and dispersion. The degree of this clustering indicates a high degree of inter-class separability and compactness of the intra-class cluster, and thus, this highlights the high discriminative ability of the framework proposed in the feature space.

Table 6: Existing CNNs, Vits, and Hybrid Models Comparison.

| 16 BS, SGD, and Data Augment | For all classes | | | | | | Individual Class accuracies | | | |
|---|---|---|---|---|---|---|---|---|---|---|
| | Accuracy | MCC | Precision | Recall | Specificity | F-1 | Glioma | Meningioma | No-tumor | Pituitary |
| **Existing CNNs** | | | | | | | | | | |
| Alexnet | 95.96 | 94.55 | 96.03 | 95.96 | 98.55 | 96.00 | 92.85 | 94.50 | 99.55 | 96.95 |
| Convnext_large | 95.88 | 94.54 | 96.16 | 95.88 | 98.52 | 96.02 | 90.05 | 96.05 | 99.81 | 97.61 |
| Densenet169 | 96.84 | 95.74 | 96.90 | 96.84 | 98.86 | 96.87 | 92.27 | 97.93 | 99.81 | 97.35 |
| Efficientnet_b4 | 92.78 | 90.20 | 92.66 | 92.78 | 97.48 | 92.72 | 92.37 | 87.21 | 96.97 | 94.57 |
| Efficientnet_v2_s | 97.07 | 95.93 | 96.89 | 97.07 | 98.95 | 96.98 | 96.42 | 94.73 | 1 | 97.13 |
| Googlenet | 96.88 | 95.70 | 96.75 | 96.89 | 98.89 | 96.82 | 95.70 | 95.20 | 99.74 | 96.91 |
| Mnasnet1_0 | 83.61 | 77.87 | 83.47 | 83.60 | 94.31 | 83.53 | 81.60 | 70.53 | 91.88 | 90.46 |
| Mobilenet_v3_small | 97.77 | 96.97 | 97.73 | 97.78 | 99.21 | 97.76 | 96.42 | 96.56 | 99.68 | 98.45 |
| Resnet50 | 96.15 | 94.76 | 96.09 | 96.15 | 98.63 | 96.12 | 96.47 | 92.57 | 98.92 | 96.64 |
| Shufflenet_v2_x2_0 | 95.51 | 94.22 | 95.75 | 95.51 | 98.58 | 95.63 | 95.79 | 94.59 | 92.69 | 98.98 |
| Squeezenet1_1 | 96.47 | 95.20 | 96.43 | 96.47 | 98.75 | 96.45 | 95.75 | 93.75 | 99.05 | 97.35 |
| Vgg13 | 96.84 | 95.71 | 96.83 | 96.85 | 98.87 | 96.84 | 93.33 | 96.99 | 99.81 | 97.26 |
| Wide_resnet50_2 | 95.94 | 94.99 | 96.48 | 95.94 | 98.77 | 96.21 | 95.60 | 97.97 | 90.86 | 99.33 |
| **Existing ViTs** | | | | | | | | | | |
| Swin_v2_s | 95.50 | 94.09 | 95.86 | 95.50 | 98.39 | 95.68 | 89.23 | 96.14 | 99.11 | 97.52 |
| Vit_b_32 | 96.96 | 95.89 | 96.99 | 96.96 | 98.91 | 96.98 | 93.77 | 96.99 | 99.74 | 97.35 |
| Vit_l_32 | 97.26 | 96.26 | 97.22 | 97.27 | 99.02 | 97.24 | 95.60 | 96.38 | 99.74 | 97.35 |
| **Hybrid and Ensemble Models** | | | | | | | | | | |
| Efficient and Vgg | 97.19 | 96.74 | 97.07 | 97.22 | 98.03 | 97.18 | 96.01 | 96.93 | 98.17 | 97.89 |
| Dens and Swin1 | 97.05 | 96.59 | 97.02 | 97.06 | 97.88 | 97.04 | 95.67 | 96.21 | 98.14 | 98.02 |
| Dens and Swin2 | 96.56 | 95.92 | 96.53 | 96.56 | 97.7 | 96.54 | 95.52 | 94.71 | 98.2 | 97.8 |
| Dense and Vit | 96.8 | 96.2 | 96.8 | 96.8 | 97.8 | 96.8 | 95.2 | 95.9 | 98.1 | 97.8 |
| **Proposed Setup** | | | | | | | | | | |
| BFS-DEF-R | 98.33 | 97.96 | 98.29 | 98.33 | 98.97 | 98.31 | 97.34 | 97.81 | 99.07 | 98.76 |
| DFE-R-BFS | 98.16 | 97.74 | 98.12 | 98.16 | 98.92 | 98.14 | 97.17 | 97.64 | 98.9 | 98.59 |
| DFE+SR-BFS | 98.13 | 97.7 | 98.11 | 98.13 | 98.91 | 98.12 | 97.14 | 97.61 | 98.87 | 98.56 |
| DFF+DR-BFS | 98.35 | 97.99 | 98.31 | 98.35 | 98.98 | 98.33 | 97.36 | 97.83 | 99.09 | 98.78 |
| BFS+DFE+DR (Proposed EDSH framework) | **98.50** | **98.05** | 98.26 | **98.50** | 98.97 | 98.38 | **97.51** | **97.98** | **99.24** | **98.93** |

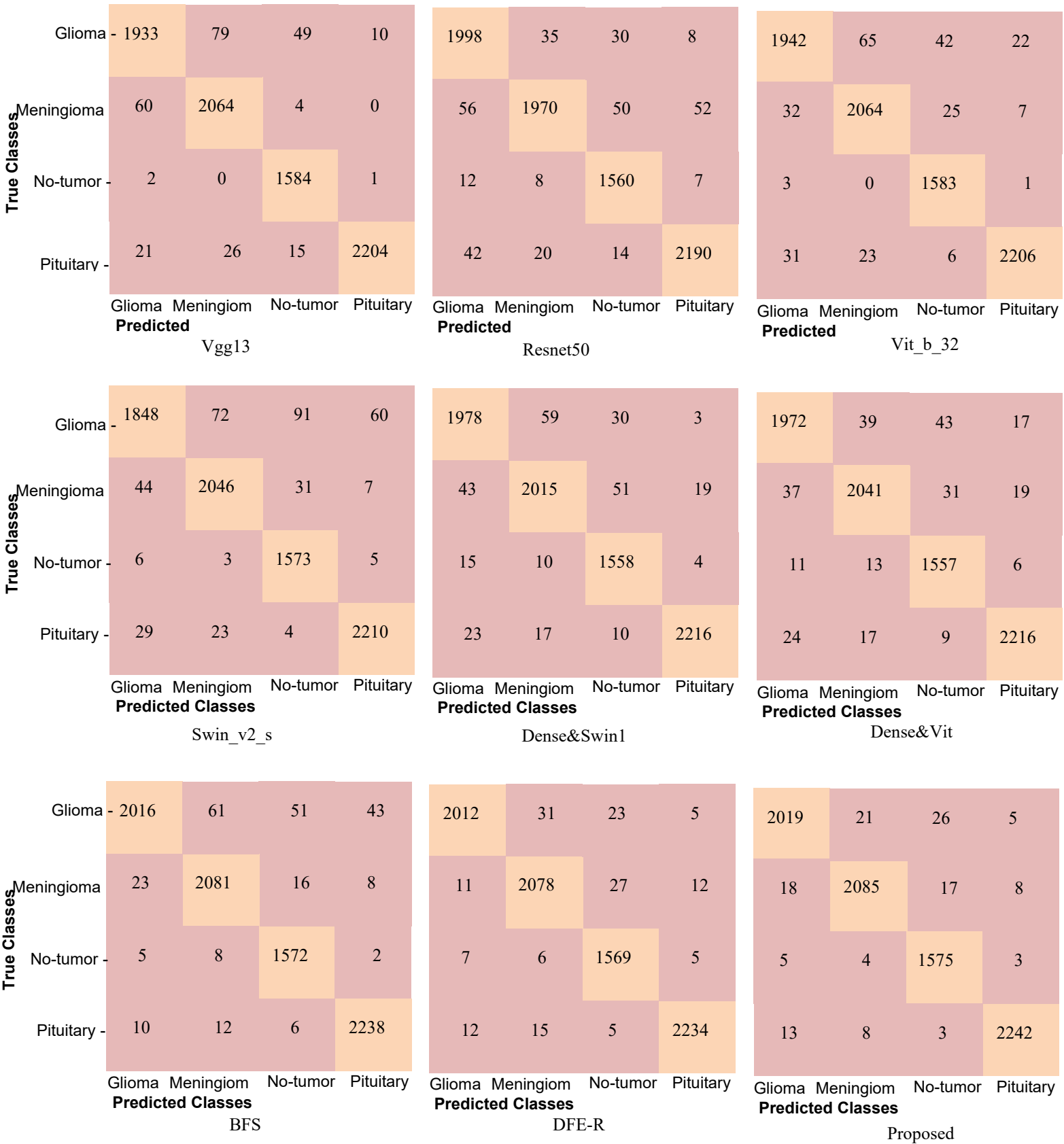

**Figure-12**: Confusion Matrix of Some existing CNN, Transformer models and Proposed (EDSH).

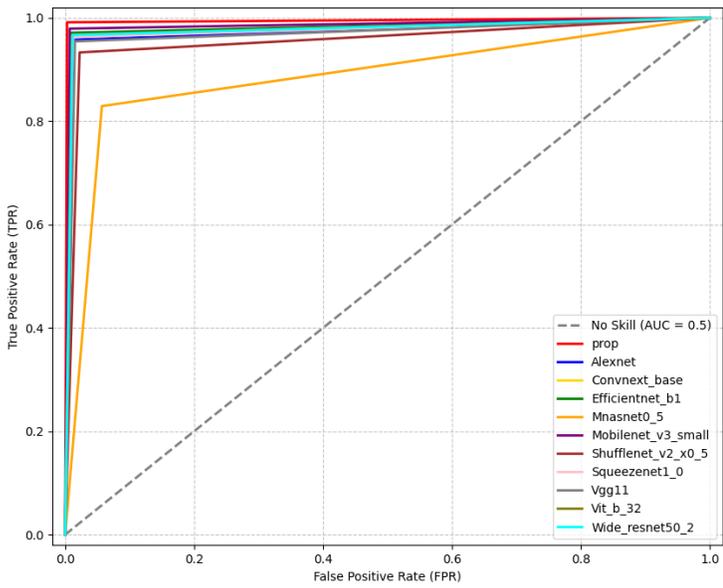
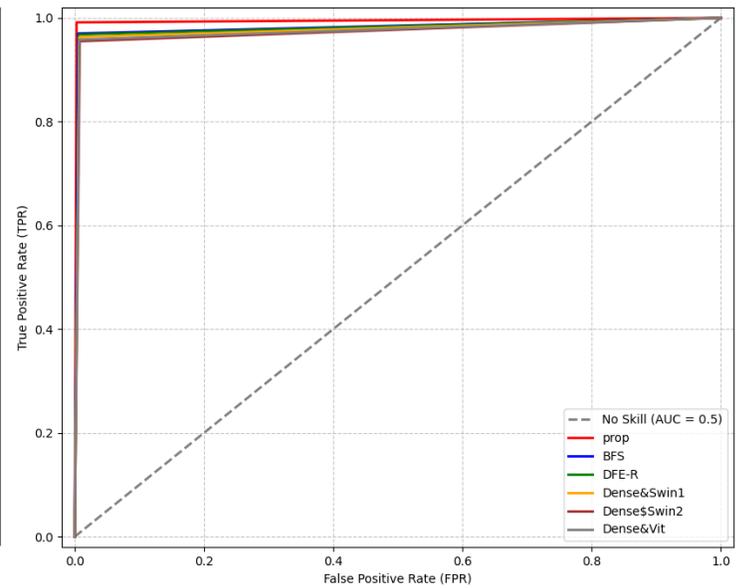
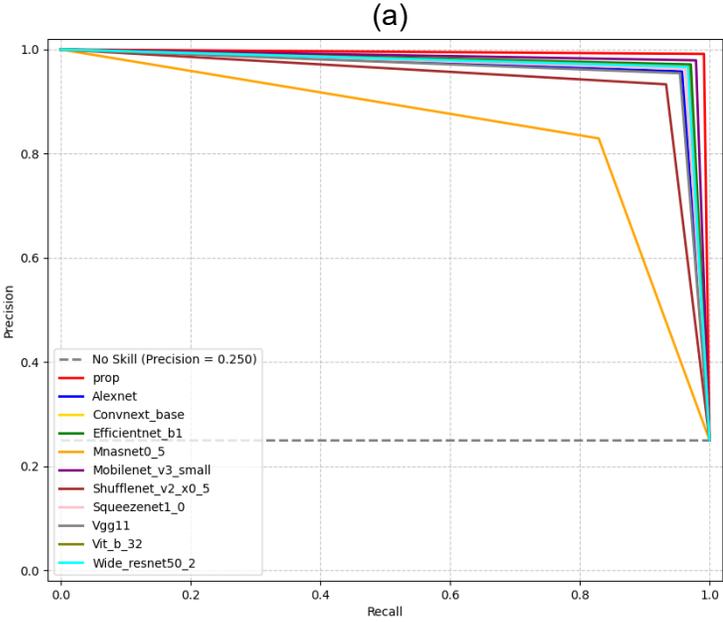
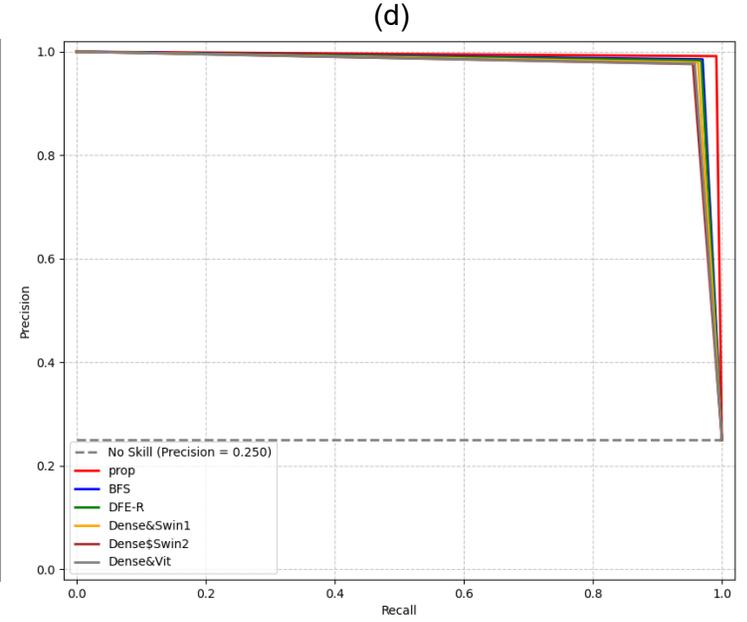
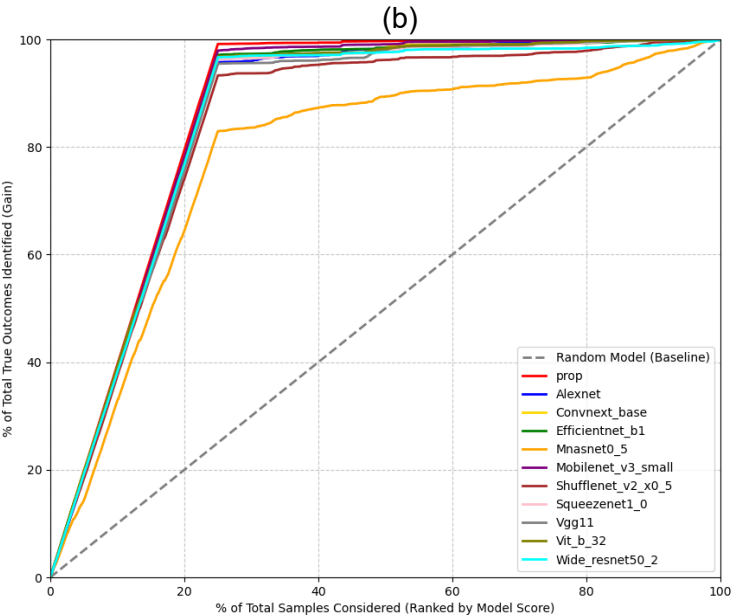
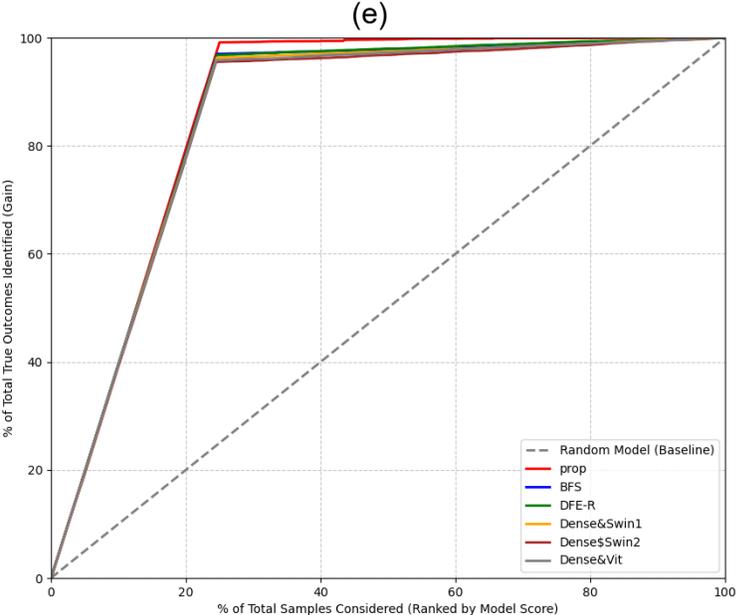

**Figure-13**: AUC, PR, Gain of Existing CNN / Vit (a)-(c) and Some Hybird models (d)-(f) compared with proposed (EDSH).

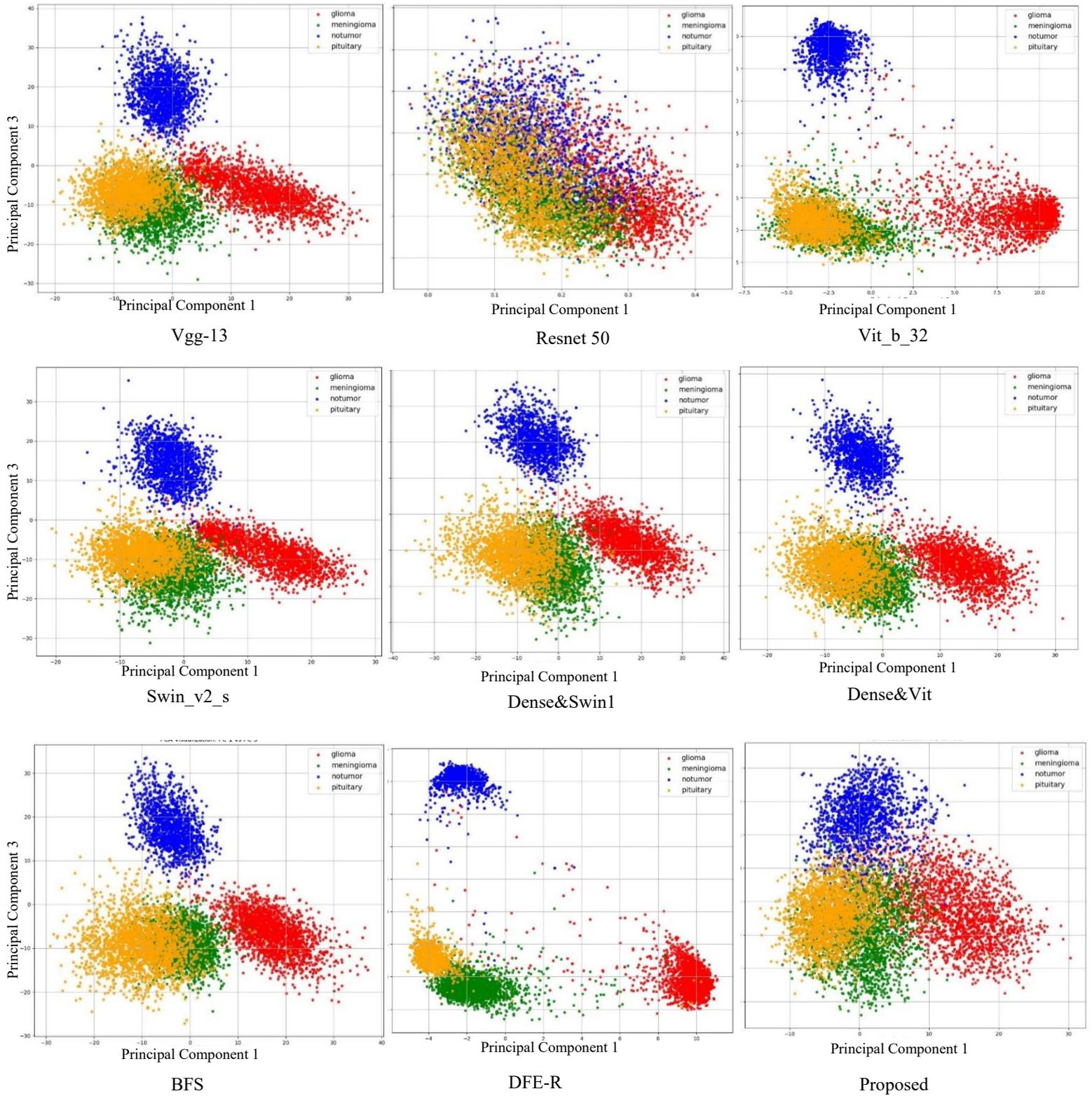

**Figure-14**: PCA analysis of some existing CNN, Vits, some hybrid models, developed models and finally proposed (EDSH).

### 5.7 Visualization Heat Maps/Bar Graphs

The authentication that proposed the EDSH framework extracted tumor-related features and focused on the right area of the MRI image; the visualization heat maps are utilized as shown in Figure-15 for all three tumor classes. The generated heatmap violet color shows the region of interest (ROI). The proposed EDSH framework, by extracting the local tumor texture, shape, and margins, then either concatenating with global relationships (BFS framework) or making contextual local features (DFE+DR framework), is able to best identify the tumor. The bar graphs of the final output tensor from the classifier are shown in Figure-16. All three tumor classes output tensor have high weightage to their corresponding tumor type, and other tensor values are at their minimum, so that can not be seen on bar graph visualization.

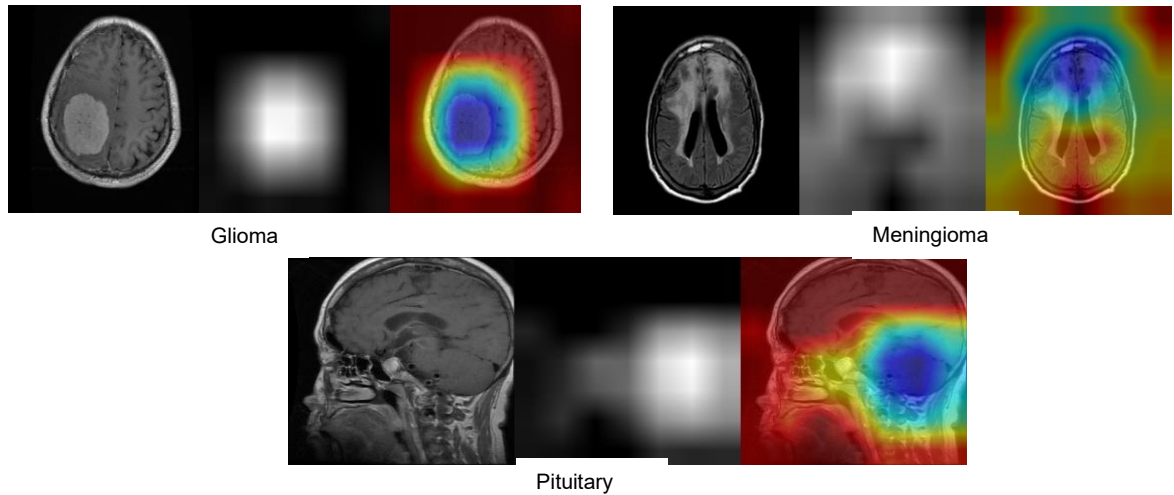

**Figure-15**: Proposed EDSH framework Heat Maps.

### 5.8 Inference time / 95% CI for recall (Comparative results)

The comparative results for recall with 95% confidence interval and inference time of existing CNNs/Vits, hybrid, ensemble, and the proposed EDSH framework are given in Table-8. The proposed EDSH framework had the lowest inference time, 348 milliseconds, as well as training time, 215 minutes, for a single epoch, as compared to other hybrid models, with the highest recall and accuracy, 98.5%. The recall evaluation metric computed with 95% confidence interval for all comparative study models, the proposed EDSH framework had the highest recall value with the lowest standard error, $98.5 \pm 0.10$.

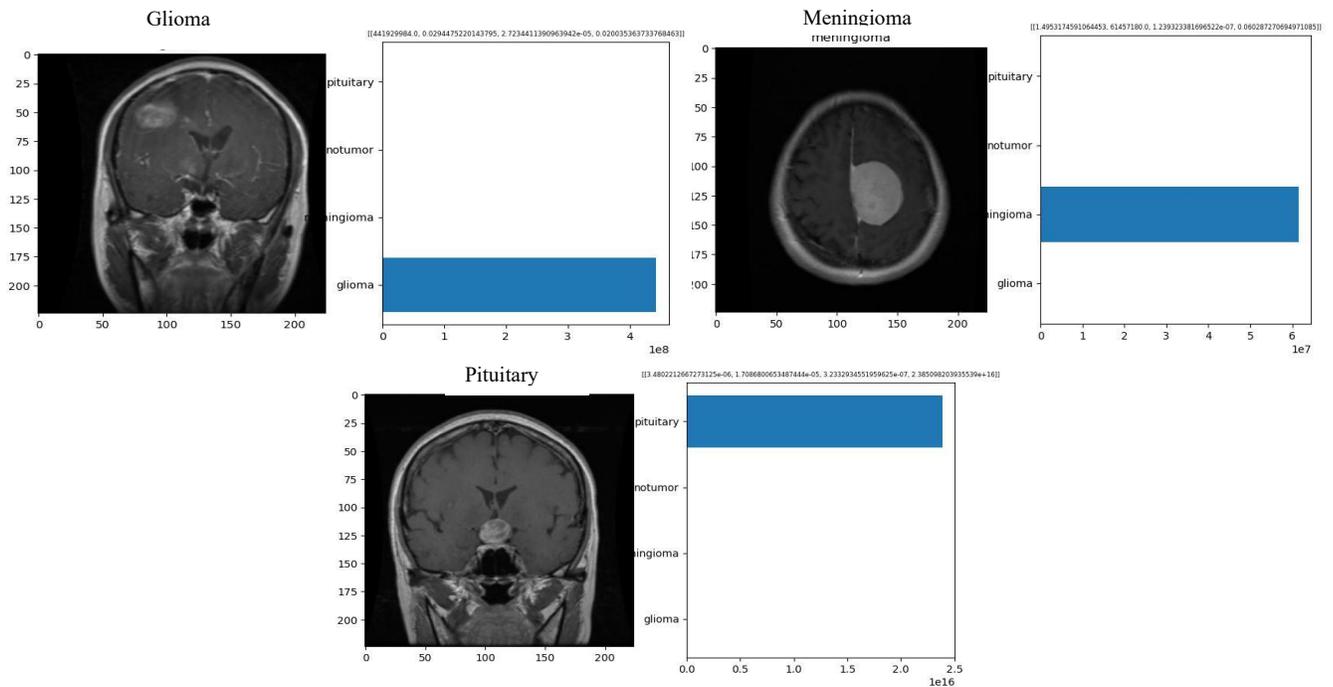

**Figure-16**: Proposed EDSA framework Visual

### 5.9 Testing on an Anonymous Dataset

To further validate the training process, anonymous (unseen) data has been collected[63] available at Kaggle, an open-source dataset which has a total of 6967 images for the classification of all four classes (recently updated). The proposed EDSH framework had final results with a 95% confidence interval for accuracy and recall are 98.66%-98.88% and 98.66%-98.89%, respectively, on this test dataset. As we further observe the details of individual accuracies (Table 7) for three tumor classes, all the results for accuracy are above 98.5%. All the detailed results are given in Table 7.

**Table 7**: Results on Unseen Dataset.

| Unseen Dataset | For all classes | | | | | | Individual Class accuracies | | | |
|---|---|---|---|---|---|---|---|---|---|---|
| | Accuracy | MCC | Precision | Recall | Specificity | F-1 | Glioma | Meningioma | Notumor | Pituitary |
| Proposed (EDSH) | 98.80 | 98.69 | 98.74 | 98.78 | 98.91 | 98.76 | 98.61 | 98.51 | 99.0 | 99.0 |

### 5.10 The Proposed EDSH Framework Significance

The significance of the proposed efficient densely swin transformer hybrid (EDSH) framework lies in the architecture innovation to achieve efficient and generalizable classification of brain tumors.

- The proposed EDSH framework utilizes two customized complementary features extracting base models having moderate complexity. The customized models' local learning includes; local textures and edges, hierarchical spatial features, and morphological features, while global learning includes: tumor relationships to other parts and structural context.

- The concatenation of local and global spaces becomes good at extracting features of high heterogeneity, complex/disordered texture, high entropy, high contrast, irregular shape, and fuzzy margins, thus misclassification of malignant tumor minimized. The concatenation preserves all the features extracted.
- The hierarchical prioritization of complementary feature spaces extracts features related to homogeneity or moderate heterogeneous structure, ordered and low entropy textures/low contrast, tumor structured having connections to other parts of the brain (dural tail or upward extension), therefore better classifying benign tumors. The hierarchical framework contextualizes the local features and minimizes the error propagation.
- The higher weightage to minimum misclassification results and their integration to the final output made an efficient and generalized decision support system for brain tumor classification.

Table 8: Existing CNNs, Vits, Hybrid, and Proposed inference time and recall results with 95% Confidence Interval

| 16 BS, SGD, Data Augment | Accuracy | Precision | Recall ± S.E. | FLOPs (G) | Inference (ms) | Time/E poch (min) |
|---|---|---|---|---|---|---|
| **Existing CNNs** | | | | | | |
| Alexnet | 95.96 | 96.03 | 95.9 ± 0.21 | 0.71 | 44.6 | 1.1 |
| Convnext_large | 95.88 | 96.16 | 95.6 ± 0.22 | 34.36 | 213.4 | 234 |
| Densenet169 | 96.84 | 96.90 | 96.9 ± 0.18 | 3.36 | 115.9 | 63.9 |
| Efficientnet_b4 | 92.78 | 92.66 | 92.8 ± 0.16 | 4.39 | 113.2 | 76.1 |
| Efficientnet_v2_s | 97.07 | 96.89 | 97.2 ± 0.18 | 8.37 | 58.8 | 43.1 |
| Googlenet | 96.88 | 96.75 | 97.0 ± 0.18 | 1.50 | 41.2 | 17.4 |
| Mnasnet1_0 | 83.61 | 83.47 | 83.2 ± 0.39 | 0.31 | 36.1 | 16.6 |
| Mobilenet_v3_small | 97.77 | 97.73 | 97.9 ± 0.15 | 0.06 | 36.7 | 8.4 |
| Resnet50 | 96.15 | 96.09 | 96.0 ± 0.21 | 4.09 | 39.8 | 44.9 |
| Shufflenet_v2_x2_0 | 95.51 | 95.75 | 95.3 ± 0.24 | 0.58 | 43.9 | 2.2 |
| Squeezenet1_1 | 96.47 | 96.43 | 96.3 ± 0.21 | 0.35 | 35.7 | 1.1 |
| Vgg13 | 96.84 | 96.83 | 96.7 ± 0.19 | 11.31 | 94.3 | 85.6 |
| Wide_resnet50_2 | 95.94 | 96.48 | 96.3 ± 0.22 | 11.40 | 106 | 81.2 |
| **Existing ViTs** | | | | | | |
| Swin_v2_s | 95.50 | 95.86 | 96.7 ± 0.22 | 11.55 | 85.9 | 98.3 |
| Vit_b_32 | 96.96 | 96.99 | 96.9 ± 0.18 | 4.41 | 60.1 | 22.4 |
| Vit_l_32 | 97.26 | 97.22 | 97.1 ± 0.18 | 15.38 | 165.4 | 72.8 |
| **Hybrid and Ensemble Models** | | | | | | |
| Efficient and vgg | 97.19 | 97.07 | 97.3 ± 0.12 | 71.78 | 366.6 | 338 |
| Dens&swin1 | 97.05 | 97.02 | 97.5 ± 0.13 | 23.14 | 407 | 263 |
| Dens&swin2 | 96.56 | 96.53 | 96.5 ± 0.15 | 49.22 | 964.4 | 333 |
| Proposed EDSH framework | **98.50** | **98.26** | **98.5 ± 0.10** | **17.56** | **347.7** | **215** |

## 6 Conclusion

The proposed EDSH framework, which combines CNN DenseNet201 and Swin Transformer Swin_t, both moderate-complexity models that exist in the PyTorch framework. It is these two models that are integrated in two separate configurations, namely Boosted Feature Space (BFS) and

Deep Feature Extraction with Double Residual Connections (DFE+DR). The BFS is good in extracting features of irregular shape, poorly defined mass, heterogeneous texture, thus better classifies glioma, while DFE+DR is good in extracting features of well-defined mass, location (outside brain), and enlargements (dural tail or upward extension), thus better classifies meningioma and pituitary tumor. Finally, both configuration classification results are given to the systematic integration (SI) module, which gives the highest weightage to the best classification results and forms the final output. The proposed EDSH framework was trained, validated, and tested on a large dataset of 40,260 MRI images of 3 tumor classes and one no-tumor class. The data set contains multimodal MRI images and all three orientations. The proposed EDSH framework recorded the best performance in accuracy as well as in recall, hence making it effective for brain tumor classification. Future works include further generalization and computational/results efficiency.

**References:**

[1] B. Aslam, A. Zafar, and U. Khalil, "Development of integrated deep learning and machine learning algorithm for the assessment of landslide hazard potential," Soft comput, vol. 25, no. 21, pp. 13493–13512, Nov. 2021, doi: 10.1007/s00500-021-06105-5.

[2] M. Mumtaz Zahoor and S. Hussain Khan, "Brain tumor MRI Classification using a Novel Deep Residual and Regional CNN."

[3] K. Fiaz et al., "Brain tumor segmentation and multiview multiscale-based radiomic model for patient's overall survival prediction," Int J Imaging Syst Technol, vol. 32, no. 3, pp. 982–999, May 2022, doi: 10.1002/ima.22678.

[4] H. Liu, G. Huo, Q. Li, X. Guan, and M. L. Tseng, "Multiscale lightweight 3D segmentation algorithm with attention mechanism: Brain tumor image segmentation," Expert Syst Appl, vol. 214, Mar. 2023, doi: 10.1016/j.eswa.2022.119166.

[5] S. Tehsin, I. M. Nasir, R. Damaševičius, and R. Maskeliūnas, "DaSAM: Disease and Spatial Attention Module-Based Explainable Model for Brain Tumor Detection," Big Data and Cognitive Computing, vol. 8, no. 9, Sep. 2024, doi: 10.3390/bdcc8090097.

[6] J. Cheng et al., "Enhanced performance of brain tumor classification via tumor region augmentation and partition," PLoS One, vol. 10, no. 10, Oct. 2015, doi: 10.1371/journal.pone.0140381.

[7] S. Deepak and P. M. Ameer, "Brain tumor classification using deep CNN features via transfer learning," Comput Biol Med, vol. 111, Aug. 2019, doi: 10.1016/j.compbiomed.2019.103345.

[8] A. Gumaei, M. M. Hassan, M. R. Hassan, A. Alelaiwi, and G. Fortino, "A Hybrid Feature Extraction Method with Regularized Extreme Learning Machine for Brain Tumor Classification," IEEE Access, vol. 7, pp. 36266–36273, 2019, doi: 10.1109/ACCESS.2019.2904145.

[9] M. Toğaçar, B. Ergen, and Z. Cömert, "BrainMRNet: Brain tumor detection using magnetic resonance images with a novel convolutional neural network model," Med Hypotheses, vol. 134, Jan. 2020, doi: 10.1016/j.mehy.2019.109531.

[10] A. Ghosh, B. Soni, and U. Baruah, "Transfer Learning-Based Deep Feature Extraction Framework Using Fine-Tuned EfficientNet B7 for Multiclass Brain Tumor Classification," Arab J Sci Eng, vol. 49, no. 9, pp. 12027–12048, Sep. 2024, doi: 10.1007/s13369-023-08607-w.

[11] N. Farajzadeh, N. Sadeghzadeh, and M. Hashemzadeh, "Brain tumor segmentation and classification on MRI via deep hybrid representation learning," Expert Syst Appl, vol. 224, Aug. 2023, doi: 10.1016/j.eswa.2023.119963.

[12] S. Pathan, P. C. Siddalingaswamy, P. Kumar, M. Pai M M, T. Ali, and U. R. Acharya, "Novel ensemble of optimized CNN and dynamic selection techniques for accurate Covid-19 screening using chest CT images," Comput Biol Med, vol. 137, Oct. 2021, doi: 10.1016/j.compbiomed.2021.104835.

[13] S. Ali et al., "Deep learning for detection and segmentation of artefact and disease instances in gastrointestinal endoscopy," Med Image Anal, vol. 70, May 2021, doi: 10.1016/j.media.2021.102002.

[14] Y. Horie et al., "Diagnostic outcomes of esophageal cancer by artificial intelligence using convolutional neural networks," Gastrointest Endosc, vol. 89, no. 1, pp. 25–32, Jan. 2019, doi: 10.1016/j.gie.2018.07.037.

[15] G. N. N. Gunesli, M. Bilal, S. E. A. E. A. Raza, and N. M. M. Rajpoot, "A Federated Learning Approach to Tumor Detection in Colon Histology Images," J Med Syst, vol. 47, no. 1, Dec. 2023, doi: 10.1007/s10916-023-01994-5.

[16] T. He et al., "Deep neural networks and kernel regression achieve comparable accuracies for functional connectivity prediction of behavior and demographics," Neuroimage, vol. 206, Feb. 2020, doi: 10.1016/j.neuroimage.2019.116276.

[17] M. Z. Asghar et al., "A Deep Neural Network Model for the Detection and Classification of Emotions from Textual Content," Complexity, vol. 2022, 2022, doi: 10.1155/2022/8221121.

[18] A. Saha et al., "A machine learning approach to radiogenomics of breast cancer: A study of 922 subjects and 529 dce-mri features," Br J Cancer, vol. 119, no. 4, pp. 508–516, Aug. 2018, doi: 10.1038/s41416-018-0185-8.

[19] H. Peng, W. Gong, C. F. Beckmann, A. Vedaldi, and S. M. Smith, "Accurate brain age prediction with lightweight deep neural networks," Med Image Anal, vol. 68, Feb. 2021, doi: 10.1016/j.media.2020.101871.

[20] D. Jha et al., "A Comprehensive Study on Colorectal Polyp Segmentation with ResUNet++, Conditional Random Field and Test-Time Augmentation," IEEE J Biomed Health Inform, vol. 25, no. 6, pp. 2029–2040, Jun. 2021, doi: 10.1109/JBHI.2021.3049304.

[21] S. Hussain Khan, A. Khan, Y. Soo Lee, M. Hassan, and W. Kyo jeong, "Segmentation of Shoulder Muscle MRI Using a New Region and Edge based Deep Auto-Encoder."

[22] Z. Zhu, E. Albadawy, A. Saha, J. Zhang, M. R. Harowicz, and M. A. Mazurowski, "Deep learning for identifying radiogenomic associations in breast cancer," Comput Biol Med, vol. 109, pp. 85–90, Jun. 2019, doi: 10.1016/j.compbiomed.2019.04.018.

[23] M. Asam et al., "IoT malware detection architecture using a novel channel boosted and squeezed CNN," Sci Rep, vol. 12, no. 1, Dec. 2022, doi: 10.1038/s41598-022-18936-9.

[24] S. I. Hassan, M. M. Alam, M. Y. I. Zia, M. Rashid, U. Illahi, and M. M. Su'ud, "Rice Crop Counting Using Aerial Imagery and GIS for the Assessment of Soil Health to Increase Crop Yield," Sensors, vol. 22, no. 21, Nov. 2022, doi: 10.3390/s22218567.

[25] J. Wang et al., "Dynamic weighted hypergraph convolutional network for brain functional connectome analysis," Med Image Anal, vol. 87, Jul. 2023, doi: 10.1016/j.media.2023.102828.

[26] D. A. Wood et al., "Deep learning models for triaging hospital head MRI examinations," Med Image Anal, vol. 78, May 2022, doi: 10.1016/j.media.2022.102391.

[27] A. Vaswani et al., "Attention Is All You Need," 2023.

[28] A. Dosovitskiy et al., "An Image is Worth 16x16 Words: Transformers for Image Recognition at Scale," Jun. 2021, [Online]. Available: http://arxiv.org/abs/2010.11929

[29] A. A. Asiri et al., "Advancing Brain Tumor Classification through Fine-Tuned Vision Transformers: A Comparative Study of Pre-Trained Models," Sensors, vol. 23, no. 18, Sep. 2023, doi: 10.3390/s23187913.

[30] S. Zheng, J. Tan, C. Jiang, and L. Li, "Automated multi-modal Transformer network (AMTNet) for 3D medical images segmentation," Phys Med Biol, vol. 68, no. 2, Jan. 2023, doi: 10.1088/1361-6560/aca74c.

[31] Y. Chen, X. Lu, and Q. Xie, "Collaborative networks of transformers and convolutional neural networks are powerful and versatile learners for accurate 3D medical image segmentation," Comput Biol Med, vol. 164, Sep. 2023, doi: 10.1016/j.compbiomed.2023.107228.

[32] F. Shamshad et al., "Transformers in medical imaging: A survey," Aug. 01, 2023, Elsevier B.V. doi: 10.1016/j.media.2023.102802.

[33] E. Goceri, "An efficient network with CNN and transformer blocks for glioma grading and brain tumor classification from MRIs," Expert Syst Appl, vol. 268, Apr. 2025, doi: 10.1016/j.eswa.2024.126290.

[34] M. Aloraini, A. Khan, S. Aladhadh, S. Habib, M. F. Alsharekh, and M. Islam, "Combining the Transformer and Convolution for Effective Brain Tumor Classification Using MRI Images," Applied Sciences (Switzerland), vol. 13, no. 6, Mar. 2023, doi: 10.3390/app13063680.

[35] K. Gao et al., "Dual-branch combination network (DCN): Towards accurate diagnosis and lesion segmentation of COVID-19 using CT images," Med Image Anal, vol. 67, Jan. 2021, doi: 10.1016/j.media.2020.101836.

[36] H. Kang et al., "Diagnosis of Coronavirus Disease 2019 (COVID-19) with Structured Latent Multi-View Representation Learning," IEEE Trans Med Imaging, vol. 39, no. 8, pp. 2606–2614, Aug. 2020, doi: 10.1109/TMI.2020.2992546.

[37] Y. Oh, S. Park, and J. C. Ye, "Deep Learning COVID-19 Features on CXR Using Limited Training Data Sets," IEEE Trans Med Imaging, vol. 39, no. 8, pp. 2688–2700, Aug. 2020, doi: 10.1109/TMI.2020.2993291.

[38] S. Hossain, A. Chakrabarty, T. R. Gadekallu, M. Alazab, and M. J. Piran, "Vision Transformers, Ensemble Model, and Transfer Learning Leveraging Explainable AI for Brain Tumor Detection and Classification," IEEE J Biomed Health Inform, vol. 28, no. 3, pp. 1261–1272, Mar. 2024, doi: 10.1109/JBHI.2023.3266614.

[39] S. Fernandes, R. Reda Mostafa, G. Peter, F. Alattar, and C.-Y. Chang, "MRI brain tumor segmentation using residual Spatial Pyramid Pooling-powered 3D U-Net." [Online]. Available: https://github.com/sanchitvj/rsppUnet-BraTS-2021.

[40] M. M. Zahoor et al., "A New Deep Hybrid Boosted and Ensemble Learning-Based Brain Tumor Analysis Using MRI," Sensors, vol. 22, no. 7, Apr. 2022, doi: 10.3390/s22072726.

[41] J. Kaur, A. K. Singh, and N. Jindal, "Systematic survey on generative adversarial networks for brain tumor segmentation and classification," Concurr Comput, vol. 35, no. 27, Dec. 2023, doi: 10.1002/cpe.7850.

[42] M. Raja and S. Vijayachitra, "A hybrid approach to segment and detect brain abnormalities from MRI scan," Expert Syst Appl, vol. 216, Apr. 2023, doi: 10.1016/j.eswa.2022.119435.

[43] C. Author, shuang Hong, and S. Hong Jin Wu Lei Zhu Weijie Chen, "PLOS ONE Brain Tumor Classification in VIT-B/16 based on Relative Position Encoding and Residual MLP Title: Brain Tumor Classification in VIT-B/16 based on Relative Position Encoding and Residual MLP Short Title: Brain Tumor Classification in VIT-B/16 based on Relative Position Encoding and Residual MLP Powered by Editorial Manager® and ProduXion Manager® from Aries Systems Corporation."

[44] J. S. Paul, A. J. Plassard, B. A. Landman, and D. Fabbri, "Deep learning for brain tumor classification," Med. Imaging 2017 Biomed. Appl. Mol. Struct. Funct. Imaging, vol. 10137, p. 1013710, Mar. 2017, doi: 10.1117/12.2254195.

[45] M. Sajjad, S. Khan, K. Muhammad, W. Wu, A. Ullah, and S. W. Baik, "Multi-grade brain tumor classification using deep CNN with extensive data augmentation," J. Comput. Sci., vol. 30, https://doi.org/10.1016/j.jocs.2018.12.003. pp. 174–182, 2019, doi:

[46] A. Çinar and M. Yildirim, "Detection of tumors on brain MRI images using the hybrid convolutional neural network architecture," Med. Hypotheses, vol. 139, pp. 109684 %@ 0306–9877, 2020.

[47] H. H. Sultan, N. M. Salem, and W. Al-Atabany, "Multi-Classification of Brain Tumor Images Using Deep Neural Network," IEEE Access, vol. 7, pp. 69215–69225, 2019.

[48] F. Özyurt, E. Sert, E. Avci, and E. Dogantekin, "Brain tumor detection based on Convolutional Neural Network with neutrosophic expert maximum fuzzy sure entropy," Measurement, vol. 147, p. 10.1016/J.MEASUREMENT.2019.07.058. 106830, Dec. 2019, doi:

[49] V. R. Sajja, "Classification of Brain tumors using Fuzzy C-means and VGG16," Turkish J. Comput. Math. Educ., vol. 12, no. 9, pp. 2103–2113, 2021.

[50] H. Mzoughi, I. Njeh, M. BenSlima, N. Farhat, and C. Mhiri, "Vision transformers (ViT) and deep convolutional neural network (D-CNN)-based models for MRI brain primary tumors images multi-classification supported by explainable artificial intelligence (XAI)," Vis. Comput., vol. 41, no. 4, pp. 2123–2142, Mar. 2025, doi: 10.1007/S00371 024-03524-X/METRICS.

[51] A. F. Al Bataineh et al., "Enhanced Magnetic Resonance Imaging-Based Brain Tumor Classification with a Hybrid Swin Transformer and ResNet50V2 Model," Applied Sciences (Switzerland), vol. 14, no. 22, Nov. 2024, doi: 10.3390/app142210154.

[52] I. Oztel, "Ensemble Deep Learning Approach for Brain Tumor Classification Using Vision Transformer and Convolutional Neural Network," Advanced Intelligent Systems, 2025, doi: 10.1002/aisy.202500393.

[53] S. Pokhrel, L. K. Dahal, N. Gupta, R. Shrestha, A. Srivastava, and A. Bhasney, "Brain Tumor Detection Application Based On Convolutional Neural Network," 2022 2nd Int. Conf. Intell. Technol. CONIT 2022, 2022, doi: 10.1109/CONIT55038.2022.9848177.

[54] N. Shamshad et al., "Enhancing Brain Tumor Classification by a Comprehensive Study on Transfer Learning Techniques and Model Efficiency Using MRI Datasets," IEEE Access, vol. 12, pp. 100407–100418, 2024, doi: 10.1109/ACCESS.2024.3430109.

[55] H. Mzoughi, I. Njeh, M. BenSlima, N. Farhat, and C. Mhiri, "Vision transformers (ViT) and deep convolutional neural network (D-CNN)-based models for MRI brain primary tumors images multi-classification supported by explainable artificial intelligence (XAI)," Vis. Comput., vol. 41, no. 4, pp. 2123–2142, Mar. 2025, doi: 10.1007/S00371-024-03524-X/METRICS.

[56] A. Khan et al., "A survey of the Vision Transformers and its CNN-Transformer based Variants."

[57] G. Huang, Z. Liu, L. van der Maaten, and K. Q. Weinberger, "Densely Connected Convolutional Networks," Jan. 2018, [Online]. Available: http://arxiv.org/abs/1608.06993

[58] P. Shaw, J. Uszkoreit, and A. Vaswani, "Self-Attention with Relative Position Representations," Apr. 2018, [Online]. Available: http://arxiv.org/abs/1803.02155

[59] Z. Liu et al., "Swin Transformer: Hierarchical Vision Transformer using Shifted Windows," Aug. 2021, [Online]. Available: http://arxiv.org/abs/2103.14030

[60] S. Tummala, S. Kadry, S. A. C. Bukhari, and H. T. Rauf, "Classification of Brain Tumor from Magnetic Resonance Imaging Using Vision Transformers Ensembling," Current Oncology, vol. 29, no. 10, pp. 7498–7511, Oct. 2022, doi: 10.3390/curroncol29100590.

[61] S. Bakheet and A. Al-Hamadi, "Automatic detection of COVID-19 using pruned GLCM-Based texture features and LDCRF classification," Comput Biol Med, vol. 137, Oct. 2021, doi: 10.1016/j.compbiomed.2021.104781.

[62] J. Wang et al., "Prior-Attention Residual Learning for More Discriminative COVID-19 Screening in CT Images," IEEE Trans Med Imaging, vol. 39, no. 8, pp. 2572–2583, Aug. 2020, doi: 10.1109/TMI.2020.2994908.

[63] https://www.kaggle.com/datasets/masumhasan/brain-tumor-mri-dataset

# Appendix A:

(a): Different CNNs/Vits models Complexity

|  | Densnet201 | Swin_t | Efficientnet_b4 | Resnet50 | Vgg13 | Googlenet | Mobilenet_v3_small | Efficientnet_v2_s | Swin_v2_s | ViT_b_32 | ViT_l_32 |
|---|---|---|---|---|---|---|---|---|---|---|---|
| **Acc on ImageNet** | **93.37** | **95.776** | 96.59 | 92.862 | 89.246 | 89.53 | 87.40 | 96.87 | 96.81 | 92.46 | 93.07 |
| **num_params (million)** | **20** | **28.2** | 19 | 25 | 133 | 6.6 | 2.5 | 21.4 | 49.7 | 88.2 | 306 |
| **GFLOPS** | **4.29** | **4.49** | 4.39 | 4.09 | 11.31 | 1.50 | 0.06 | 8.37 | 11.55 | 4.41 | 15.38 |
| **File size (MB)** | **77.4 MB** | **108.2** | 74.5 | 97.8 | 507.5 | 49.7 | 9.8 MB | 82.7 | 190.7 | 336.6 | 1169.4 |